\documentclass[twoside,11pt]{article}    
\usepackage{tr2e}

\begin{document}

\newtheorem{method}{Method}[section]
\newtheorem{procedure}{Procedure}[section]
\newtheorem{observation}{Observation}[section]

\renewcommand\floatpagefraction{1.0}

\def\odt{{\textstyle{1\over 2}}}
\def\maxarg{\mathop{\rm maxarg}}          
\def\minarg{\mathop{\rm minarg}}          

\trheading{IDSIA-12-02, Version 2.0}{arXiv:cs.AI/0207097 v2}{1-43}{December 23, 2002}{}{J. Schmidhuber} 
\ShortHeadings{Optimal Ordered Problem Solver}{Schmidhuber}
\firstpageno{1}

\title{Optimal Ordered Problem Solver} 
\author{\name J\"{u}rgen Schmidhuber  \email 
juergen@idsia.ch - www.idsia.ch/\~{ }juergen
\\
\addr IDSIA, Galleria 2, 6928 Manno-Lugano, Switzerland}

\editor{}

\maketitle
 
\begin{abstract}

In a quite pragmatic sense {\sc oops} is the fastest general way of
solving one task after another, always optimally exploiting solutions
to earlier tasks when possible. It can be used for increasingly hard
problems of optimization or prediction.  Suppose there is only one
task and a bias in form of a probability distribution $P$ on programs
for a universal computer.  In the $i$-th phase $(i=1,2,3, \ldots)$
of asymptotically optimal {\em non}incremental universal 
search (Levin, 1973, 1984) we test
all programs $p$ with runtime $\leq 2^iP(p)$ until the task is solved.
Now suppose there is a sequence of tasks, e.g., the $n$-th task is to
find a shorter path through a maze than the best found so far. To reduce
the search time for new tasks, previous {\em incremental} extensions
of universal search tried to modify $P$ through
experience with earlier tasks---but in a heuristic and non-general and
suboptimal way prone to overfitting.  {\sc Oops}, however, does it right.

Tested self-delimiting program prefixes (beginnings of code that may
continue) are immediately executed while being generated. They grow
by one instruction whenever they request this.  The storage for the
first found program computing a solution to the current task becomes
non-writeable. Programs tested during search for solutions to later task
may copy non-writeable code into separate modifiable storage, to edit
it and execute the modified result.  Prefixes may also recompute the
probability distribution on their suffixes in arbitrary computable ways.
To solve the $n$-th task we sacrifice half the total search time for
testing (via universal search) programs that have the most recent
successful program as a prefix.  The other half remains for testing
fresh programs starting at the address right above the top non-writeable
address. When we are searching for a universal solver for all tasks
in the sequence we have to time-share the second half (but not the
first!) among all tasks $1..n$.  For realistic limited computers we
need efficient backtracking in program space to reset storage contents
modified by tested programs.  We introduce a recursive procedure
for doing this in time-optimal fashion.

{\sc Oops} can solve tasks unsolvable by traditional reinforcement
learners and AI planners, such as {\em Towers of Hanoi} with 30 disks
(minimal solution size $> 10^9$).  In our experiments OOPS demonstrates
incremental learning by reusing previous solutions to discover a prefix
that temporarily rewrites the distribution on its suffixes, such that
universal search is accelerated by a factor of 1000. This illustrates
how {\sc oops} can benefit from self-improvement and metasearching,
that is, searching for faster search procedures.

We mention several {\sc oops} variants and outline {\sc oops}-based
reinforcement learners.  Since {\sc oops} will scale to larger problems in
essentially unbeatable fashion, we also examine its physical limitations.

\end{abstract}

\vspace{1cm}

\newpage
\noindent
{\em Based on 
arXiv:cs.AI/0207097 v1 (TR-IDSIA-12-02 version 1.0, July 2002)
\citep{Schmidhuber:02oops,Schmidhuber:02nips}.
All sections are illustrated by Figures \ref{storage}
and \ref{searchtree}
at the end of this paper. Frequently used symbols are collected
in reference Table \ref{general} (general {\sc oops}-related symbols) 
and Table \ref{specific} (less important implementation-specific symbols,
explained in the appendix, Section \ref{language}).
}

\begin{small}
\tableofcontents
\end{small}

\section{Introduction}
\label{intro}

We train children and most machine learning systems
on sequences of harder and harder tasks.
This makes sense since new problems often are more easily 
solved by reusing or adapting solutions to previous problems.

Often new tasks depend on solutions for earlier tasks.
For example, given an NP-hard optimization problem, the 
$n$-th task in a sequence of tasks may be to find an 
approximation to the unknown optimal 
solution such that the new approximation is at 
least 1 \% better (according to some measurable 
performance criterion) than the best found so far. 

Alternatively we may want to find
a strategy for solving {\em all} tasks in a given sequence of
more and more complex tasks. 
For example, we might want to teach our learner a program that
computes {\sc fac}$(n)= 1 \times 2 \times \ldots n$ for any given positive
integer $n$.
Naturally, the $n$-th task  in the ``training sequence''
will be to compute  {\sc fac}$(n)$.

In general we would like our learner to continually 
profit from useful information conveyed by solutions
to earlier tasks. To do this in an optimal fashion,
the learner may also have to improve the way 
it exploits earlier solutions. 
Is there a general yet time-optimal way of achieving such 
a feat?  Indeed, there is.  The Optimal Ordered Problem Solver
({\sc oops}) is a simple, general, 
theoretically sound way of solving one task 
after another, efficiently searching the space of programs that compute 
solution candidates, including programs that
organize and manage and adapt and reuse earlier acquired knowledge.

\subsection{Overview}
\label{overview}

Section \ref{survey}
will survey previous relevant work on general optimal
search algorithms.
Section \ref{oops} will use the framework of universal computers
to explain {\sc oops} and how it benefits from
incrementally extracting useful knowledge hidden in training sequences.
The remainder of the paper is devoted to ``Realistic'' {\sc oops}  
which uses a recursive procedure for time-optimal planning
and backtracking
in program space 
to perform efficient storage management 
(Section \ref{realistic}) 
on realistic, limited computers.
Appendix \ref{language}
describes an pilot implementation of Realistic {\sc oops} based on
a stack-based universal programming language inspired
by {\sc Forth} \citep{Forth:70},
with initial primitives for defining and calling recursive
functions, iterative loops, arithmetic operations, domain-specific
behavior, and even for rewriting the search procedure itself.
Experiments in Section \ref{experiments} use
the language of Appendix \ref{language} to solve 60 tasks in a row:
we first teach {\sc oops} something about recursion, by training it to
construct samples of the simple context free language $\{ 1^k2^k \}$ 
($k$ 1's followed by $k$ 2's), for $k$ up to 30.
This takes roughly 0.3 days on a standard personal computer (PC).
Thereafter, within a few additional days, 
{\sc oops} demonstrates the benefits of incremental knowledge transfer:
it exploits certain properties of its previously 
discovered universal $1^k2^k$-solver
to greatly accelerate the search for a universal solver
for all $k$ disk {\em Towers of Hanoi} problems, solving all
instances up to $k=30$ (solution size $2^k-1$).
Previous, less general reinforcement learners
and {\em non}learning AI planners 
tend to fail for much smaller instances.

\section{Survey of Universal Search 
and Suboptimal Incremental Extensions}
\label{survey}

Let us start by briefly reviewing
general, asymptotically optimal 
search methods by \cite{Levin:73,Levin:84} and \cite{Hutter:01fast}.
These methods are {\em non}incremental in the sense that they
do not attempt to accelerate the search for solutions
to new problems through experience with previous problems. 
We will point out drawbacks of existing heuristic extensions
for {\em incremental} search.
The remainder of the paper will describe  {\sc oops} which
overcomes these drawbacks.

\subsection{Bias-Optimality}
\label{bias}

For the purposes of this paper, a problem $r$ is defined by a
recursive procedure $f_r$ that takes as an input any potential solution
(a finite symbol string $y \in Y$, where $Y$ represents
a search space of solution candidates) 
and outputs 1 if $y$ is a solution to $r$,
and 0 otherwise.  Typically the goal is to find 
as quickly as possible some $y$ that solves $r$.  

Define a probability distribution $P$ on a finite or infinite
set of programs for a given computer.  $P$ represents the searcher's initial bias
(e.g., $P$ could be based on program length, or on a probabilistic syntax diagram).
A {\em bias-optimal} searcher will not spend more
time on any solution candidate than it deserves,
namely, not more than the candidate's probability times the total search time:
\begin{definition} [{\sc Bias-Optimal Searchers}]
\label{bias-optimal}
Given is a problem class $\cal R$,
a search space $\cal C$ of solution candidates
(where  any problem $r \in \cal R$ should have a solution in $\cal C$),
a task-dependent bias in form of conditional probability
distributions $P(q \mid r)$ on the candidates $q \in \cal C$,
and a predefined procedure that creates and tests any given $q$
on any $r \in \cal R$ within time $t(q,r)$ (typically unknown in advance).
A searcher is {\em $n$-bias-optimal} ($n \geq 1$) if  
for any maximal total search time $T_{max} > 0$
it is guaranteed to solve any problem $r \in \cal R$
if it has a solution $p \in \cal C$
satisfying $t(p,r) \leq P(p \mid r)~T_{max}/n$.    
It is {\em bias-optimal} if  $n=1$.      
\end{definition}
This definition makes intuitive sense: the most probable candidates
should get the lion's share of the total search time, in a way that
precisely reflects the initial bias.

\subsection{Near-Bias-Optimal Nonincremental Universal Search}
\label{levin}

The following straight-forward method
(sometimes referred to as {\em Levin Search} or {\sc Lsearch})
is near-bias-optimal.
For simplicity, we notationally suppress conditional dependencies on the current problem.
Compare
\cite{Levin:73,Levin:84,Solomonoff:86,Schmidhuber:97bias,LiVitanyi:97,Hutter:01fast} (Levin also attributes similar ideas to Allender):
\begin{method}[{\sc Lsearch}]
\label{lsearch}
Set current time limit T=1.  {\sc While} problem not solved {\sc do:}
\begin{quote}
Test all programs $q$ such that $t(q)$,
the maximal time spent on creating and running and testing $q$,
satisfies $t(q) < P(q)~T$.  Set $T := 2 T.$      
\end{quote}
\end{method}
Note that 
{\sc Lsearch}  has the optimal order of computational complexity:
Given some problem class,
if some unknown optimal program $p$ requires $f(k)$ steps to solve a
problem instance of size $k$, 
then {\sc Lsearch} 
will need at most $O(P(p) f(k)) = O(f(k))$ steps ---
the constant factor  $P(p)$ may be huge but does not depend on $k$.

The near-bias-optimality of  {\sc Lsearch}    
is hardly affected by the fact that for each value of $T$ we
repeat certain computations for the previous value.     
Roughly half the total search time is still spent on $T$'s maximal
value (ignoring hardware-specific overhead for parallelization and
nonessential speed-ups due to halting programs
if there are any).
Note also that the time for testing is properly taken into account here:
any result whose validity is hard to test is automatically penalized.

Universal
{\sc Lsearch}
provides inspiration for
nonuniversal but very practical methods which are optimal
with respect to a limited search space, while suffering
only from very small slowdown factors.
For example, designers of planning procedures often just
face a binary choice between two options such as
depth-first and breadth-first search.
The latter is often preferrable, but its greater demand for
storage may eventually require to move data from on-chip memory to disk.
This can slow down the search by a factor of 10,000 or more.
A straightforward solution in the spirit of {\sc Lsearch} is to       
start with a 50 \% bias towards either technique, and use both
depth-first and breadth-first search in parallel --- this will
cause a slowdown factor of at most 2 with respect to the best      
of the two options (ignoring a bit of overhead for parallelization).   
Such methods have presumably been used long before Levin's 1973 paper.
\cite{Wiering:96levin} and \cite{Schmidhuber:97bias} used
rather general but nonuniversal
variants of {\sc Lsearch}  
to solve machine learning toy problems unsolvable by traditional
methods.  Probabilistic alternatives
based on {\em probabilistically chosen maximal program runtimes}   
in {\em Speed-Prior} style \citep{Schmidhuber:00v2,Schmidhuber:02colt}
also outperformed traditional methods on toy problems
\citep{Schmidhuber:95kol,Schmidhuber:97nn}.
\nocite{Schmidhuber:02ijfcs}

\subsection{Asymptotically Fastest Nonincremental Problem Solver}
\label{hutter}

Recently my postdoc \cite{Hutter:01fast} developed 
a more complex asymptotically
optimal search algorithm for {\em all} well-defined problems.
{\sc Hsearch} (or {\em Hutter Search}) cleverly allocates part of the total
search time to searching the space of proofs for provably correct
candidate programs with provable upper runtime bounds;
at any given time it
focuses resources on those programs with the currently
best proven time bounds.
Unexpectedly, {\sc Hsearch} manages to reduce
the constant slowdown factor to a value smaller than $5$.
In fact, it can be made smaller than $1 + \epsilon$, where $\epsilon$
is an arbitrary positive constant (M. Hutter, personal communication, 2002).

Unfortunately, however,  {\sc Hsearch} is not yet the final word
in computer science, since the search in proof space
introduces an unknown {\em additive} problem class-specific
constant slowdown, which again may be huge.
While additive constants generally are preferrable to
multiplicative ones, both types may make universal
search methods practically infeasible---in the real world 
constants do matter. For example, the last to cross the finish 
line in the Olympic 100 m dash may be only a constant factor 
slower than the winner, but this will not comfort him.
And since constants beyond $2^{500}$ do not even make sense within 
this universe,  both {\sc Lsearch} and {\sc Hsearch} may be viewed 
as academic exercises demonstrating that the $O()$ notation 
can sometimes be practically irrelevant despite its 
wide use in theoretical computer science.

\subsection{Previous Work on Incremental Extensions of Universal Search}
\label{incremental}

\begin{small}
\hspace{5.6cm}
{\sl ``Only math nerds would consider  $2^{500}$ finite.''} 
 {\sc (Leonid Levin)}
\vspace{0.2cm}
\end{small}

{\sc Hsearch}  and {\sc Lsearch} (Sections \ref{levin},
\ref{hutter})  neglect one potential source of speed-up: they are
nonincremental in the sense that they do not attempt to minimize their 
constant slowdowns
by exploiting experience collected in previous searches for solutions to
earlier tasks. They simply ignore
the constants --- from an asymptotic point of view, incremental search does 
not buy anything.  

A heuristic attempt 
\citep{Schmidhuber:97bias} 
to greatly reduce the constants through experience
was called {\em Adaptive} {\sc Lsearch} or {\sc Als} 
--- compare related ideas by
\cite{Solomonoff:86,Solomonoff:89}.
Essentially {\sc Als} works as follows: whenever {\sc Lsearch}
finds a program $q$ that computes a solution for the current problem,
$q$'s probability $P(q)$ is 
substantially increased using a ``learning rate,''
while probabilities of alternative programs decrease
appropriately.
Subsequent  {\sc Lsearch}es for new problems then use the adjusted
$P$, etc.  
\cite{Schmidhuber:97bias} and \cite{Wiering:96levin}
used a nonuniversal variant of this approach to solve 
reinforcement learning (RL) tasks 
in partially observable environments
unsolvable by traditional RL
algorithms.

Each  {\sc Lsearch} invoked by {\sc Als} is bias-optimal with respect to 
the most recent adjustment of $P$.
On the other hand, the rather arbitrary $P$-modifications themselves 
are not necessarily optimal. They might
lead to {\em overfitting} in the following sense: modifications of $P$ after the
discovery of a solution to problem 1 could actually be harmful and slow
down the search for a solution to problem 2, etc.
This may provoke a loss of near-bias-optimality with respect to the initial
bias during exposure to subsequent tasks. Furthermore, {\sc Als} has a fixed
prewired method for changing $P$ and cannot improve this method by experience.
The main contribution of this paper is 
to overcome all such drawbacks in a principled way.

\subsection{Other Work on Incremental Learning}
\label{other}

Since the early attempts of \cite{Newell:63} at building a ``General Problem
Solver'' ---see also \cite{SOAR:93}---much work has been done to develop
mostly heuristic machine learning
algorithms that solve new problems based on experience with
previous problems, by incrementally shifting the
inductive bias in the sense of \cite{Utgoff:86}.
Many pointers to
{\em learning by chunking,
learning by macros,
hierarchical learning,
learning by analogy,}
etc.  can be found in the book
by \cite{Mitchell:97}.
Relatively recent general attempts include
program evolvers such as
{\sc Adate} \citep{Olsson:95} and simpler heuristics
such as {\em Genetic Programming (GP)} \citep{Cramer:85,Banzhaf:98}.
Unlike logic-based program synthesizers \citep{Green:69,Waldinger:69,Deville:94},
program evolvers use biology-inspired concepts of
{\em Evolutionary Computation} \citep{Rechenberg:71,Schwefel:74}
or {\em Genetic Algorithms} \citep{Holland:75}
to evolve better and better computer programs.
Most existing GP implementations, however, do not even allow for
programs with loops and recursion, thus ignoring a main motivation for
search in program space.
They either have very limited search spaces (where solution candidate runtime
is not even an issue), or are far from bias-optimal, or both.
Similarly, traditional reinforcement learners \citep{Kaelbling:96}
are neither general nor close to being bias-optimal.

A first step to make GP-like methods bias-optimal would be
to allocate runtime to tested programs in proportion
to the probabilities of the mutations or ``crossover operations''
that generated them. Even then there would still be
room for improvement, however, since GP has quite limited ways of making
new programs from previous ones---it does not learn better
program-making strategies.

This brings us to several previous publications on
{\em learning to learn} or {\em metalearning} \citep{Schmidhuber:87},
where the goal is to learn better learning algorithms
through self-improvement without human intervention---compare
the human-assisted self-improver by \cite{Lenat:83}.
We introduced the concept of 
incremental search for improved,
probabilistically generated code that modifies
the probability distribution on the possible code continuations:
{\em incremental self-improvers} \citep{Schmidhuber:97ssa}
use the {\em success-story algorithm} SSA
to undo those self-generated probability modifications
that in the long run do not contribute to increasing the learner's
cumulative reward per time interval.
An earlier meta-GP algorithm
\citep{Schmidhuber:87}
was designed to learn better GP-like strategies;
\cite{Schmidhuber:87}
also combined principles of reinforcement
learning economies \citep{Holland:85}
with a ``self-referential'' metalearning approach.
A gradient-based metalearning technique
\citep{Schmidhuber:93selfreficann}
for {\em continuous} program spaces of differentiable
recurrent neural networks (RNNs) was also designed to favor
better learning algorithms;
compare the remarkable recent success of the related but
technically improved RNN-based metalearner by \cite{Hochreiter:01meta}.
 
The algorithms above generally are not near-bias-optimal though.
The method discussed in this paper, however,
combines optimal search and incremental self-improvement,
and will be {\em $n$-bias-optimal}, where $n$ is a small and practically
acceptable number, such as 8.

\section{OOPS on Universal Computers}
\label{oops}

An informed reader familiar with concepts such as universal computers 
\citep{Turing:36} and
self-delimiting programs 
\citep{Levin:74,Chaitin:75} 
will probably understand the simple basic principles 
of {\sc oops} by just reading the abstract. For the others,
Subsection \ref{basics} will start the formal description
of {\sc oops} by introducing notation and explaining program
sets that are prefix codes.  
Subsection \ref{principles} will provide {\sc oops} pseudocode and
point out its essential properties
and a few essential differences to previous work.
The remainder of the paper is about practical implementations
of the basic principles on realistic computers with limited storage.

\begin{table*}
\label{general}
\begin{center}
\begin{tabular}{|c|c|c|c|c|c|c|}    \hline
{\bf Symbol} & {\bf Description} \\  \hline
$Q$ & variable set of instructions or tokens \\  \hline
$Q_i$ & $i$-th possible token (an integer) \\  \hline
$n_Q$ & current number of tokens \\  \hline
$Q^*$ & set of strings over alphabet $Q$, containing  the search space of programs \\  \hline
$q$ & total current code $\in Q^*$ \\  \hline
$q_n$ & $n$-th token of code $q$ \\  \hline
$q^n$ & $n$-th frozen program $\in Q^*$, where total code $q$ starts with $q^1q^2 \ldots$ \\  \hline
$qp$ & {\em $q$-pointer} to the highest address of code $q=q_{1:qp}$ \\  \hline
$a_{last}$ & start address of a program (prefix) solving all tasks so far \\  \hline
$a_{frozen}$ & top frozen address, can only grow, $1 \leq a_{last} \leq  a_{frozen} \leq qp$  \\  \hline
$q_{1:a_{frozen}}$ & current code bias   \\  \hline
$R$ & variable set of tasks, ordered in cyclic fashion; each task has a computation tape \\  \hline
$S$ & set of possible tape symbols (here: integers) \\  \hline
$S^*$ & set of strings over alphabet $S$, defining possible states stored on tapes \\  \hline
$s^i$ & an element of $S^*$ \\  \hline
$s(r)$ & variable state of task $r \in R$, stored on tape $r$ \\  \hline
$s_i(r)$ & $i$-th component of $s(r)$ \\  \hline
$l(s)$ & length of any string $s$  \\  \hline
$z(i)(r)$ & equal to $q_i$ if $0 < i \leq l(q)$ or equal to $s_{-i}(r)$ if $-l(s(r)) \leq i \leq 0$ \\  \hline
$ip(r)$ & current instruction pointer of task $r$, encoded on tape $r$ within state $s(r)$ \\  \hline
$p(r)$ & variable probability distribution on  $Q$, encoded on tape $r$ as part of $s(r)$ \\  \hline
$p_i(r)$ & current history-dependent probability of selecting $Q_i$ if $ip(r)=qp+1$ \\  \hline
\end{tabular}
\end{center}
\caption{{\em
Symbols used to explain the basic principles of {\sc oops} (Section \ref{oops}).
}}
\end{table*}

\subsection{Formal Setup and Notation}
\label{basics}
Unless stated otherwise or obvious,
to simplify notation,
throughout the paper newly introduced variables
are assumed to be integer-valued and to cover the range implicit in the context.
Given some finite or countably infinite alphabet $Q=\{Q_1,Q_2, \ldots \}$,
let $Q^*$ denote the set of finite sequences or strings over $Q$,
where $\lambda$ is the empty string. 
Then let $q,q^1,q^2, \ldots \in Q^*$ be (possibly variable) strings.
$l(q)$ denotes the number of symbols in string $q$, where
$l(\lambda) = 0$;
$q_n$ is the $n$-th symbol of string $q$;
$q_{m:n}= \lambda$ if $m>n$ and $q_m q_{m+1} \ldots q_n$
otherwise
(where $q_0 := q_{0:0} := \lambda$).
$q^1q^2$ is the concatenation of $q^1$ and $q^2$ (e.g.,
if $q^1=abc$ and $q^2=dac$ then $q^1q^2 = abcdac$).

Consider countable alphabets $S$ and $Q$.
Strings $s,s^1,s^2, \ldots \in S^*$ represent possible internal
{\em states} of a computer; strings $q,q^1,q^2, \ldots \in Q^*$
represent {\em token sequences} or {\em code} or {\em programs} for manipulating states.
We focus on $S$ being the set of integers
and $Q := \{ 1, 2, \ldots, n_Q \}$ representing a set of
$n_Q$ instructions of some universal
programming language \citep{Goedel:31,Turing:36}.
(The first universal programming language
due to \cite{Goedel:31}
was based on integers as well, but ours will be more practical.)
$Q$ and $n_Q$ may be variable: new tokens
may be defined by combining previous tokens,
just as traditional
programming languages allow for the declaration of new tokens
representing new procedures.
Since  $Q^* \subset S^*$,
substrings within states may also encode programs.

$R$ is a set of currently unsolved tasks.
Let the variable $s(r) \in S^*$ denote the current state of task $r \in R$,
with $i$-th component $s_i(r)$ on a {\em computation tape} $r$
(a separate tape holding a separate state for each task, initialized
with task-specific inputs represented by the initial state).
Since subsequences on tapes may also represent executable code,
for convenience we combine 
current code $q$ and any given current state $s(r)$ 
in a single {\em address space},
introducing negative and positive addresses ranging
from $-l(s(r))$ to $l(q)+1$,
defining the content of address $i$ as
$z(i)(r) := q_i$ if $0 < i \leq l(q)$
and $z(i)(r) := s_{-i}(r)$ if $-l(s(r)) \leq i \leq 0$.
All dynamic task-specific data
will be represented at nonpositive addresses (one code, many tasks).
In particular, the current instruction pointer
{\em ip(r)} $:= z(a_{ip}(r))(r)$ of task $r$
(where $ip(r) \in {-l(s(r)), \ldots, l(q)+1})$
can be found at (possibly variable)
address $a_{ip}(r) \leq 0$.
Furthermore, $s(r)$ also encodes
a modifiable probability distribution
$p(r) = \{ p_1(r), p_2(r), \ldots, p_{n_Q}(r) \}$
$(\sum_i p_i(r) = 1)$ on $Q$. 

Code is executed in a way inspired by self-delimiting binary programs
\citep{Levin:74,Chaitin:75} 
studied in the theory of Kolmogorov complexity and
algorithmic probability \citep{Solomonoff:64,Kolmogorov:65}.
Section \ref{try} will present
details of a practically useful variant of this approach.
Code execution is time-shared sequentially among all current tasks.
Whenever any $ip(r)$ has been initialized or changed 
such that its new value points to a valid address 
$\geq -l(s(r))$ but $\leq l(q)$, 
and this address contains some executable token $Q_i$, 
then  $Q_i$ will define task $r$'s next instruction to be 
executed. 
The execution may change $s(r)$ including $ip(r)$.
Whenever the time-sharing process works on task $r$
and $ip(r)$ points to the smallest positive currently 
unused address $l(q)+1$, $q$ will grow by one token
(so $l(q)$ will increase by 1), and the current value of $p_i(r)$ 
will define the current probability of selecting $Q_i$ as the next token,
to be stored at new address $l(q)$ and to be executed immediately. 
That is, executed program beginnings or {\em prefixes}
define the probabilities of their possible suffixes.
(Programs will be interrupted through errors or halt instructions or 
when all current tasks are solved or when certain search time limits 
are reached---see Section \ref{principles}.)

To summarize and exemplify: programs are grown
incrementally, token by token; their 
prefixes are immediately executed while being created;
this may modify some task-specific internal state
or memory, and may transfer control back to previously
selected tokens (e.g., loops).
To add a new token to some program
prefix, we first have to wait until the
execution of the prefix so far {\em explicitly
requests} such a prolongation, by setting an
appropriate signal in the internal state.
Prefixes that cease to request any further tokens
are called self-delimiting programs or simply
programs  (programs are their own prefixes).
So our procedure yields {\em task-specific
prefix codes} on program space: with any given task,
programs that halt because they have found a solution
or encountered some error cannot request any more tokens.
Given a single task and the current task-specific 
inputs, no program can be the prefix of
another one.  On a different task, however, the same program
may continue to request additional tokens.

$a_{frozen} \geq 0 $ is a variable address that can increase but not decrease.
Once chosen, the {\em code bias} $q_{0:a_{frozen}}$
will remain unchangeable forever --- it is a (possibly empty)
sequence of programs $q^1q^2 \ldots$, some of them
prewired by the user, others {\em frozen} after previous successful
searches for solutions to previous task sets (possibly completely
unrelated to the current task set $R$).

To allow for programs that exploit previous solutions,
the instruction set $Q$ should contain instructions
for invoking or calling code found below $a_{frozen}$, 
for copying such code
into some $s(r)$, and for editing the copies and executing the results.
Examples of such instructions will be given in 
the appendix (Section \ref{language}).

\subsection{Basic Principles of OOPS}
\label{principles}

Given a sequence of tasks, we solve one task after another in the given order. 
The solver of the $n$-th task $(n \geq 1)$ will be a program $q^i$ $(i \leq n)$
stored such that it occupies successive addresses somewhere between 1 and $l(q)$.
The solver of the $1$st task will start at address 1.
The solver of the $n$-th task $(n>1)$ will either start
at the same address as the solver of the
$n-1$-th task, or right after its end address.
To find a universal solver for all tasks in a given task sequence, do: 
\begin{method}[{\sc oops}]
\label{oopsalgbasics}

{\bf FOR} task index $n=1,2,\ldots$ {\bf DO:}

\noindent 
{\bf 1.} 
Initialize current time limit $T := 2$.

\noindent 
{\bf 2.} 
Spend at most $T/2$ on a variant of 
{\sc Lsearch} that searches for a program 
solving task $n$ and starting
at the start address $a_{last}$ of the most 
recent successful code (1 if there is none).
That is, the problem-solving program 
either must be equal to
$q_{a_{last}:a_{frozen}}$ or must have 
$q_{a_{last}:a_{frozen}}$ as a prefix.
If solution found, go to {\bf 5.} 

\noindent
{\bf 3.}
Spend at most $T/2$ on {\sc Lsearch} for a fresh
program that starts at the first writeable address 
and solves {\em all} tasks $1..n$.
If solution found, go to {\bf 5.} 

\noindent
{\bf 4.} Set $T := 2 T$, and go to {\bf 2.}

\noindent
{\bf 5.} 
Let the top non-writeable address $a_{frozen}$ point to the end of the 
just discovered
problem-solving program.

\end{method}
\subsection{Essential Properties of OOPS}
\label{properties}
The following observations highlight important
aspects of {\sc oops} and point out in which sense {\sc oops}
is optimal.

\begin{observation}
A program starting at $a_{last}$ and solving task $n$ will also solve
all tasks up to $n$.
\end{observation}
{\em Proof} (exploits the nature of self-delimiting programs):
Obvious for $n=1$. For $n> 1$: By induction, the code 
between $a_{last}$ and $a_{frozen}$,
which cannot be altered any more, already solves all tasks up to $n-1$.
During its application to task $n$
it cannot request any additional tokens that could harm its performance
on these previous tasks. So those of its prolongations that solve task $n$
will also solve tasks $1, \ldots, n-1$.

\begin{observation}
$a_{last}$ does not increase if task $n$ can be
more quickly solved by testing prolongations of $q_{a_{last}:a_{frozen}}$
on task $n$, than by testing fresh programs starting above
$a_{frozen}$ on all tasks up to $n$.
\end{observation}

\begin{observation}
Once we have found an optimal solver for all
tasks in the sequence, at most half of the total future 
time will be wasted on searching for faster alternatives.
\end{observation}

\begin{observation}
Unlike the learning rate-based
bias shifts of {\sc Als} (Section \ref{incremental}),
those of {\sc oops} do not reduce the probabilities of programs that
were meaningful and executable {\em before} the addition of  any new $q^i$.
\end{observation}
But consider formerly meaningless program prefixes trying to access
code for earlier solutions when there weren't any:
such prefixes may suddenly become prolongable and successful,
once some solutions to earlier tasks have been stored.
That is, unlike with {\sc Als}
the acceleration potential of {\sc oops} is not bought at
the risk of an unknown slowdown due to nonoptimal changes of the
underlying probability distribution through a heuristically
chosen learning rate. As new tasks come along, {\sc oops}
remains near-bias-optimal with respect to the initial bias,
while still being able to profit in from subsequent code
bias shifts in an optimal way.

\begin{observation}
\label{fastest}
Given the initial bias and subsequent
code bias shifts due to $q^1, q^2, \ldots, $ no bias-optimal
searcher with the same initial bias will solve the current task set
substantially faster than {\sc oops}.  
\end{observation}
Ignoring hardware-specific overhead 
(e.g., for measuring time and switching between 
tasks), {\sc oops} will lose at most a factor 2 through
allocating half the search time to prolongations of
$q_{a_{last}:a_{frozen}}$, and another factor 2 through the incremental
doubling of time limits in {\sc Lsearch} (necessary because 
we do not know in advance the final time limit). 

\begin{observation}
\label{startaddress}
If the current task (say, task $n$)
can already be solved by some previously frozen program $q^k$,
then the probability of a solver for task $n$ is at least equal to 
the probability of the most probable program computing the start 
address $a(q^k)$ of $q^k$ and setting 
instruction pointer $ip(n):=a(q^k)$.
\end{observation}

\begin{observation}
As we solve more and more tasks, thus collecting and freezing
more and more $q^i$,
it will generally become harder and harder to identify and address
and copy-edit useful code segments within the earlier solutions.
\end{observation}
As a consequence we
expect that much of the knowledge embodied by 
certain $q^j$ actually
will be about how to access and copy-edit 
and otherwise use programs $q^i$ ($i<j$)
previously stored below  $q^j$.

\begin{observation}
Tested program prefixes may rewrite the probability distribution 
on their suffixes in
computable ways (based on previously frozen $q^i$), thus 
temporarily redefining the 
search space structure of {\sc Lsearch},
essentially rewriting the search procedure.
If this type of metasearching for faster search algorithms is 
useful to accelerate the search
for a solution to the current problem, 
then {\sc oops} will automatically exploit this.
\end{observation}
Since there is no fundamental difference between
domain-specific problem-solving programs and programs
that manipulate probability distributions and rewrite
the search procedure itself, we collapse both learning and
metalearning in the same time-optimal framework.

\begin{observation}
If the overall goal is just to solve one task after another, as opposed to
finding a universal solver for all tasks, it suffices to test 
only on task $n$ in step {\bf 3.}
\end{observation}
For example, in an optimization context the $n$-th task usually is not
to find a solver for all tasks in the sequence, but just to find
an approximation to some unknown optimal solution such that the
new approximation is better than the best found so far.

\subsection{Summary} 

{\sc Lsearch} is about optimal time-sharing, given one problem.
{\sc Oops} is about optimal time-sharing, given a sequence of problems.
The basic principles of {\sc Lsearch} can be explained in one line:
time-share all program tests such that each program gets a constant
fraction of the total search time.  Those of {\sc oops} require just a
few more lines: use self-delimiting programs and freeze those that were
successful; given a new task, spend a fixed fraction of the total search
time on programs starting with the most recently frozen prefix (test only
on the new task, never on previous tasks); spend the rest of the time
on fresh programs
(when looking for a universal solver, test them on all previous tasks).

{\sc Oops} spends part of the total search time for a new problem on
programs that exploit previous solutions in computable ways.  If the
new problem can be solved faster by copy-editing/invoking previous
code than by solving the new problem from scratch, then  {\sc oops}
will find this out.  If not, then at least it will not suffer from the
previous solutions.

If {\sc oops} is so simple indeed, then why does the paper not end here
but has 31 additional pages? The answer is: to describe the additional
efforts required to make OOPS work on realistic limited computers,
as opposed to universal machines.

\section{OOPS on Realistic Computers}
\label{realistic}

Unlike the Turing machines originally 
used to describe {\sc Lsearch} and {\sc Hsearch},
realistic computers have limited storage. So we need to efficiently reset
storage modifications computed by the numerous programs {\sc oops} is testing.
Furthermore, our programs typically will be composed from more complex 
primitive instructions than those of typical Turing machines. In what 
follows we will address such issues in detail.

\begin{figure}[hbt]
\centerline{\psfig{figure=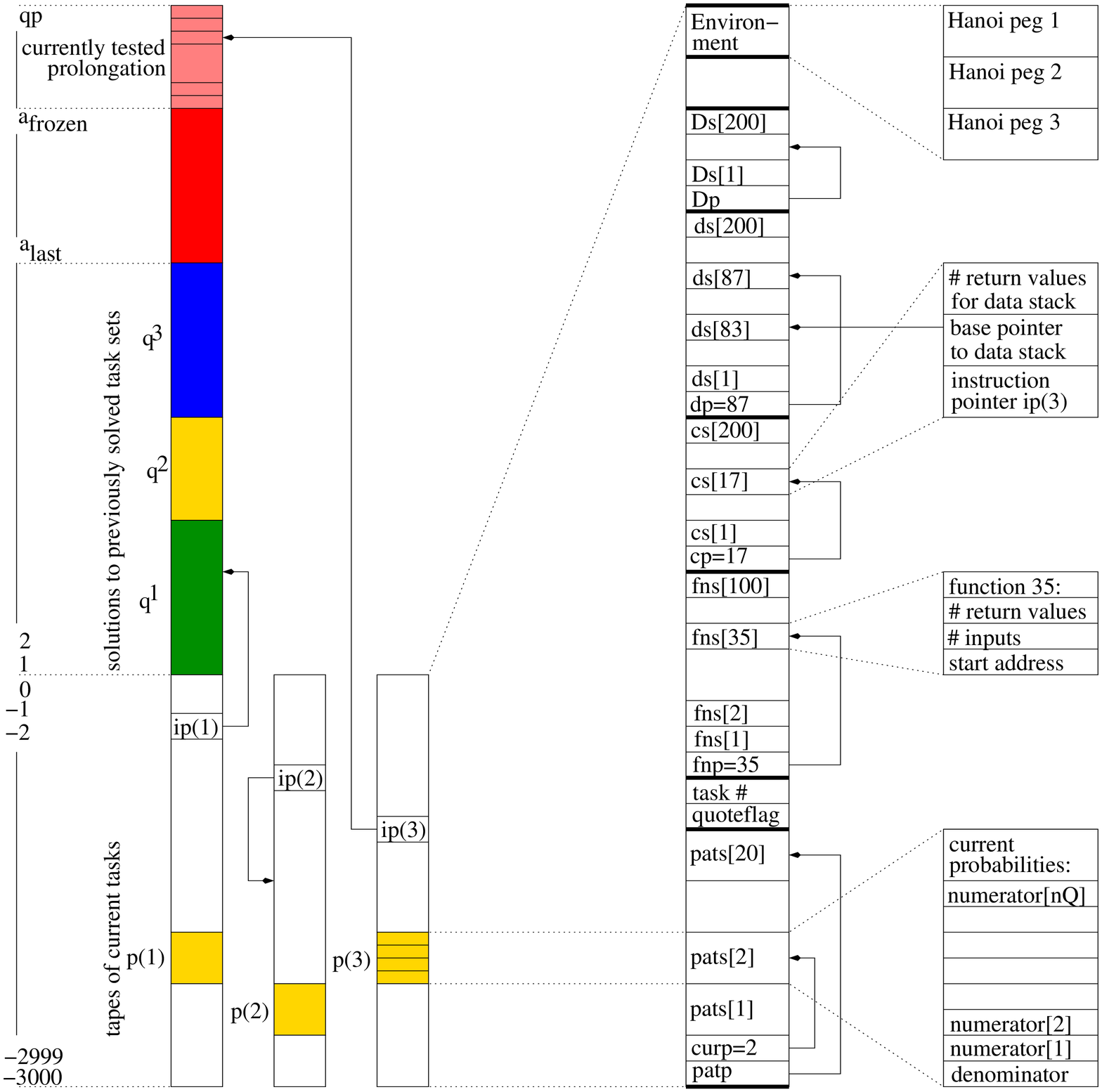,angle=0,height=16cm}}
\caption{
{\small
Storage snapshot during an {\sc oops} application.
{\bf Left:} general picture (Section \ref{oops}). {\bf Right:} language-specific
details for a particular
{\sc Forth}-like programming language (Sections \ref{language} and \ref{experiments}).
{\bf Left:} 
current code $q$ ranges from addresses $1$ to $qp$ and includes
previously frozen programs $q^1,q^2,q^3$.
Three unsolved tasks require three tapes (lower left) with addresses $-3000$ to $0$.
Instruction pointers $ip(1)$ and $ip(3)$ point to code in $q$, $ip(2)$ to code
on the 2nd tape. 
Once, say, $ip(3)$ points right above the topmost address $qp$,
the probability of the next instruction (at $qp+1$) 
is determined by the current probability
distribution $p(3)$ on the possible tokens.
{\sc oops} spends equal time on programs
starting with prefix $q_{a_{last}:a_{frozen}}$ (tested only on the most recent task, since such programs
solve all previous tasks, by induction), 
and on all programs starting at $a_{frozen}+1$ (tested on all tasks). 
{\bf Right:} Details of a single tape.
There is space for several alternative self-made probability distributions on $Q$,
each represented by $n_Q$ numerators and a common denominator.
The pointer {\em curp} 
\hspace{1.7cm}
determines 
which distribution to use for the next token request.
There is a stack {\em fns} of self-made function definitions, each with a pointer to
its code's start address, and its numbers of input arguments and return values expected 
on data stack {\em ds} (with stack pointer {\em dp}).  The 
dynamic runtime
stack {\em cs} 
handles all function calls. 
Its top 
entry holds the current instruction pointer {\em ip} and 
the current base pointer into {\em ds} below the arguments of the most recent
call. There is also space for an auxiliary stack {\em Ds}, and for 
representing modifiable aspects of the environment.  
}
}
\label{storage}
\end{figure}

\begin{figure}[hbt]
\centerline{\psfig{figure=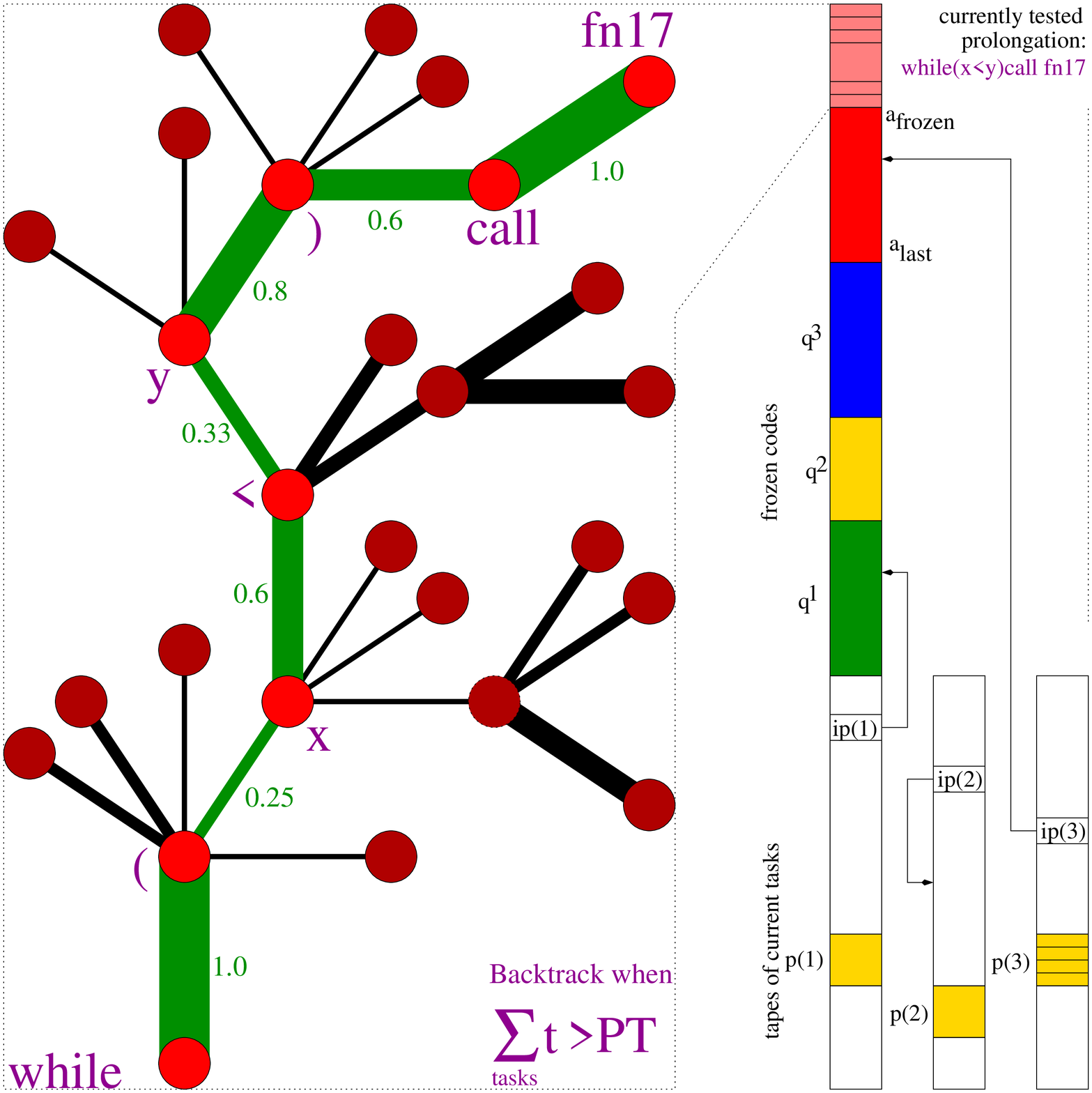,angle=0,height=16cm}}
\caption{
{\small
Search tree during an {\sc oops} application; compare Section \ref{try} and 
Figure \ref{storage}. 
The tree branches are program prefixes (a single prefix may modify several 
task-specific tapes in parallel);
nodes represent tokens; widths of connections between nodes stand for temporary,
task-specific transition
probabilities encoded on the modifiable tapes. Prefixes
may contain (or call previously frozen) 
subprograms that dynamically modify the conditional 
probabilities during runtime, thus rewriting the suffix search procedure.  
In the example, the currently tested prefix (above the previously frozen codes) 
consists of the token sequence {\em while $(x<y)$ call fn17}
(real values denote transition probabilities).
Here {\em fn17} might be a time-consuming primitive, say,
for manipulating the arm of a realistic virtual robot.
Before requesting an additional token, this prefix may run
for a long time, thus changing many components of numerous tapes.
Node-oriented backtracking  through procedure {\bf Try} will
restore partially solved task sets and
modifications of internal states and continuation probabilities 
once 
\hspace{0.8cm}
there is an error or the sum of the runtimes of the current 
prefix on all current tasks exceeds the prefix probability multiplied 
by the 
total search time so far.
See text for details.
}
}
\label{searchtree}
\end{figure}

\subsection{Multitasking \& Prefix 
Tracking By Recursive Procedure {\bf ``Try''}}
\label{try}

{\sc Hsearch} and {\sc Lsearch}
assume potentially infinite storage. Hence they
may largely ignore questions of storage management.
In any practical system, however, we have to efficiently reuse
limited storage. Therefore, in both subsearches of 
Method \ref{oopsalgbasics} (steps {\bf 2} and {\bf 3}), 
Realistic {\sc oops} evaluates alternative
prefix continuations 
by a practical, token-oriented
backtracking procedure that can deal with several tasks in parallel,
given some {\em code bias} in the form of previously found code.

The novel recursive method {\bf Try} below 
essentially conducts a depth-first search in program space,
where the branches of the search tree are program prefixes
(each modifying a bunch of task-specific states),
and backtracking (partial resets of partially solved task sets and
modifications of internal states and continuation probabilities) 
is triggered once the sum of the runtimes of the current prefix 
on all current tasks exceeds the 
current time limit 
multiplied by the prefix probability 
(the product of the history-dependent
probabilities of the previously selected prefix components in $Q$).
This ensures near-{\em bias-optimality} (Def. \ref{bias-optimal}),
given some initial probabilistic bias on program space $ \subseteq Q^*$.

Given task set $R$, the current goal is to solve all tasks $r \in R$,
by a single program that either appropriately uses or extends
the current code $q_{0:a_{frozen}}$
(no additional freezing will take place before all tasks in $R$ are solved).

\subsubsection{Overview of ``Try''}
\label{tryoverview}

We assume an initial set of user-defined primitive behaviors
reflecting prior knowledge and assumptions of the user. 
Primitives may be assembler-like instructions or
time-consuming software, such as, say, theorem provers, 
or matrix operators for neural network-like parallel architectures,  
or trajectory generators for robot simulations, 
or state update procedures for multiagent systems,
etc.  Each primitive is represented by a token $\in Q$.
It is essential that those primitives 
whose runtimes are not known in advance
can be interrupted by {\sc oops} at any time.

The searcher's initial bias is also affected
by initial, user-defined, task-dependent probability 
distributions on the finite or infinite search space of 
possible self-delimiting program prefixes. In the simplest case we start
with a maximum entropy distribution on the tokens, and
define prefix probabilities as the products of the 
probabilities of their tokens.
But prefix continuation probabilities may 
also depend on previous tokens in 
context sensitive fashion defined by a probabilistic
syntax diagram.  In fact, we even permit that any executed prefix 
assigns a task-dependent, self-computed
probability distribution to its own possible suffixes
(compare Section \ref{basics}).

Consider the left-hand side of Figure \ref{storage}.
All instruction pointers $ip(r)$ of all
current tasks $r$ are initialized by some address,
typically below the topmost code address, 
thus accessing the code bias common to all tasks, 
and/or using task-specific code fragments written
into tapes.
All tasks keep executing their instructions in parallel
until interrupted or all tasks are solved, 
or until some task's instruction pointer points to the yet unused
address right after the 
topmost code address.  The latter case 
is interpreted as a request for code prolongation through a new token,
where each token has a probability 
according to the present task's current state-encoded 
distribution on the possible next tokens.
The deterministic method {\bf Try} systematically 
examines all possible code extensions in a depth-first fashion
(probabilities of prefixes are just used to order them for runtime
allocation).
Interrupts and backtracking to previously selected tokens
(with yet untested alternatives) and the corresponding partial resets
of states and task sets
take place whenever one of the tasks encounters an error,
or the product of the task-dependent probabilities of the currently selected tokens
multiplied by the {\em sum} of the runtimes on all tasks exceeds a given
total search time limit $T$. 

To allow for efficient backtracking,
{\bf Try} tracks effects of tested program prefixes,
such as task-specific state modifications (including probability distribution changes)
and partially solved task sets, to reset
conditions for subsequent tests of alternative, yet untested
prefix continuations in an optimally efficient fashion
(at most as expensive as the prefix tests themselves).

Since programs are created online while they are being executed,
{\bf Try} will never create impossible
programs that halt before all their tokens are read. 
No program that halts on a given task can be 
the prefix of another program halting on the same task. 
It is important to see, however, that
in our setup a given prefix that has solved one task 
(to be removed from the current task set)
may continue to demand tokens as it tries to solve other tasks.

\subsubsection{Details of ``Try:'' Bias-Optimal
Depth-First Planning in Program Space}
\label{trydetails}

To allow us to efficiently undo state changes, we
use global Boolean
variables $mark_i(r)$ (initially {\sc False})
for all possible state components $s_i(r)$.
We initialize time
$t_0 := 0;$ probability $P := 1$;
{\em q-pointer}
$qp := a_{frozen}$
and state $s(r)$ --- including
$ip(r)$ and $p(r)$ --- with task-specific information
for all task names $r$ in a so-called {\em ring} $R_0$  of 
tasks, where the expression {\em ``ring''}
indicates that the tasks are ordered in cyclic fashion;
$\mid R \mid$ denotes the number of tasks in ring $R$.
Given a global search time limit $T$,
we {\bf Try}
to solve all tasks in $R_0$, by
using existing code in $q= q_{1:qp}$ and / or by
discovering an appropriate prolongation of $q$:
\begin{center}
-----------------------------------------------------------------------------------------
\end{center}
\begin{method}[{\sc Boolean} {\bf Try ($qp, r_0, R_0, t_0, P$)}]
{\em ($r_0 \in R_0$; returns {\sc True} or {\sc False}; may have the side effect of
increasing $a_{frozen}$ and thus prolonging the 
frozen code $q_{1:a_{frozen}}$):}

\vspace{0.3cm}
 \noindent
 {\bf 1.}
 Make an empty stack $\cal S$;
 set local variables $r := r_0; R := R_0; t:= t_0;$
 {\em Done}$:=$ {\sc False}.

\noindent
{\sc While} there are unsolved tasks
 ($\mid R \mid > 0$)
   {\sc and} there is enough time left ($t \leq PT$)
      {\sc and} instruction pointer valid ($-l(s(r)) \leq ip(r) \leq qp$)
          {\sc and} instruction valid ($1 \leq z(ip(r))(r) \leq n_Q$)
	       {\sc and} no halt condition 
	       is encountered
	             {\em (e.g., error such as division by 0,
		     or robot bumps into obstacle; 
		     evaluate conditions in the above order until
		            first satisfied, if any)}
			            {\sc Do:}
\begin{quote}
Interpret / execute token $z(ip(r))(r)$ according to the rules
of the given programming language, continually increasing
$t$ by the consumed time.  This may
modify $s(r)$ including instruction pointer $ip(r)$ and distribution
$p(r)$, but not code $q$.
Whenever the execution changes some
state component $s_i(r)$ whose $mark_i(r)=$ {\sc False},
set $mark_i(r) :=$ {\sc True} and
save the previous value $\hat{s}_i(r)$ by
pushing the triple $(i, r, \hat{s}_i(r))$
onto $\cal S$.
Remove $r$ from  $R$ if solved.
{\sc If} $\mid R \mid > 0$,
{\bf set $r$ equal to the next task in ring $R$} {\em (like in
the round-robin method of standard operating systems).}
{\sc Else} set {\em Done} $:=$  {\sc True};
$a_{frozen} := qp$
{\em (all tasks solved; new code frozen, if any).}
\end{quote}

\noindent
{\bf 2.}
Use $\cal S$ to efficiently reset only 
the modified $mark_i(k)$ to {\sc False}
{\em (the global mark variables will be needed again in step {\bf 3}),}
but do not pop $\cal S$ yet.
 
\vspace{0.3cm}
 \noindent
 {\bf 3.}
 {\sc If} $ip(r)=qp+1$ ({\bf i.e., if there is an online request for
 prolongation of the current prefix through a new token}):
 {\sc While} {\em Done} $=$ {\sc False} and there  is some
 yet untested token $Z \in Q$ (untried since $t_0$ as
 value for $q_{qp+1}$) {\sc Do:}
\begin{quote}
 Set $q_{qp+1}:=Z$ and
 {\em Done} $:=$ {\bf Try ($qp+1, r, R, t, P * p(r)(Z)$)},
 where $p(r)(Z)$ is $Z$'s probability
 according to current distribution $p(r)$.
\end{quote}

\vspace{0.3cm}
\noindent
{\bf 4.}
Use $\cal S$ to efficiently
restore only those $s_i(k)$ changed since $t_0$,
thus restoring all tapes to their states at the
beginning of the current invocation of {\bf Try}.
This will also restore instruction pointer $ip(r_0)$ and
original search distribution $p(r_0)$.
Return the value of {\em Done}.
\end{method}
\begin{center}
-----------------------------------------------------------------------------------------
\end{center}
A successful {\bf Try} will solve all tasks, possibly
increasing $a_{frozen}$ and prolonging total code $q$. In any case
{\bf Try} will completely restore all states of all tasks.
It never wastes time on
recomputing previously computed results of prefixes,
or on restoring unmodified state components and marks,
or on already solved tasks
--- tracking / undoing effects of prefixes
essentially does not cost more than their execution.
So the $n$ in Def. \ref{bias-optimal}
of {\em $n$-bias-optimality}
is not greatly affected by the undoing procedure: 
we lose at most a factor 2, ignoring hardware-specific overhead 
such as the costs of single {\em push} and {\em pop} 
operations on a given computer, or the costs of measuring time,
etc.

Since the distributions $p(r)$ are modifiable, we speak of 
self-generated continuation probabilities.
As the variable
suffix $q' := q_{a_{frozen}+1 : qp}$ of the 
total code  $q = q_{1:qp}$ is growing, 
its probability can be readily updated:
\begin{equation}
\label{prob}
P(q' \mid s^0)= \prod_{i=a_{frozen}+1}^{qp}  P^i(q_i \mid s^i),
\end{equation}
where $s^0$ is an initial state,
and $P^i(q_i \mid s^i)$ is the probability of $q_i$, 
given the state $s^i$ of the task $r$ 
whose variable distribution $p(r)$ (as a part of $s^i$) 
was used to determine the probability
of  token $q_i$ at the moment it was selected. 
So we allow 
the probability of $q_{qp+1}$ to depend on $q_{0:qp}$ and 
intial state $s^0$ in a fairly arbitrary computable fashion.
Note that unlike the
traditional Turing machine-based setup
by \cite{Levin:74} and \cite{Chaitin:75}
(always yielding binary programs $q$ with probability $2^{-l(q)}$)
this framework of self-generated continuation probabilities
allows for token selection probabilities close to 1.0,
that is, even long programs may have high probability.

{\em Example.} In many programming languages the probability of
token ``('', given a previous token ``{\sc While}'', equals 1.
Having observed the ``('' there is
not a lot of new code to execute yet --- in such cases
the rules of the programming language will typically demand
another increment of instruction pointer {\em ip(r)}, which 
will lead to the request
of another token through subsequent increment of the topmost code address.
However, once we have observed a complete
expression of the form ``{\sc While} (condition) {\sc Do} (action),''
it may take a long time until the conditional loop
--- interpreted via $ip(r)$ --- is exited
and the top address is incremented again, thus asking for a new token.

The {\em round robin} {\bf Try} variant above keeps circling 
through all unsolved tasks, executing one instruction at a time.
Alternative {\bf Try} variants 
could also sequentially work on each task
until it is solved, then try to prolong the resulting $q$
on the next task, and so on, appropriately restoring previous tasks once
it turns out that the current task cannot be solved 
through prolongation of the prefix solving 
the earlier tasks.
One potential advantage of  {\em round robin} {\bf Try} is that
it will quickly discover whether 
the currently studied prefix 
causes an error for at least one task, in which case it
can be discarded immediately.

{\bf Nonrecursive C-Code}.
An efficient iterative (nonrecursive)
version of {\bf Try} for a broad variety of initial
programming languages was implemented in C.
Instead of local stacks $\cal S$,
a single global stack is used to save and restore
old contents of modified cells of all tapes / tasks.

\subsection{Realistic OOPS for Finding Universal Solvers}
\label{unisolve}

Recall that the instruction set $Q$ should contain instructions
for invoking or calling code found below $a_{frozen}$, for copying such code
into $s(r)$, and for editing the copies and executing the results
(examples in Appendix \ref{language}).

Now suppose there is an ordered sequence of tasks $r_1, r_2, \ldots$.
Inductively suppose we have solved the first $n$ tasks
through programs stored below address $a_{frozen}$,
and that the most recently discovered program
starting at address $a_{last} \leq a_{frozen}$
actually solves all of them, possibly using
information conveyed by earlier programs $q^1,q^2,\ldots$.
To find a program solving the first $n+1$ tasks, Realistic
{\sc oops} 
invokes {\bf Try} as follows (using set notation for task rings,
where the tasks are ordered in cyclic fashion---compare basic 
Method \ref{oopsalgbasics}):

\begin{center}
-----------------------------------------------------------------------------------------
\end{center}
\begin{method}[Realistic {\sc oops} (n+1)] 
\label{oopsalg}
Initialize current time 
limit $T := 2$ and  $q$-pointer $qp := a_{frozen}$ 
{\em (top frozen address)}.

\vspace{0.2cm}
\noindent
{\bf 1.}
 Set instruction pointer $ip(r_{n+1}) := a_{last}$
 {\em (start address of code solving all tasks up to $n$)}.
\begin{quote}
  {\sc If} {\bf Try ($qp, r_{n+1}, \{ r_{n+1} \}, 0, \frac{1}{2}$)} then exit.

  {\em (This means that half the search time is assigned
  to the most recent $q_{a_{last}:a_{frozen}}$
  and all possible prolongations thereof).}
\end{quote}

\noindent
{\bf 2.}
{\sc If} it is possible to initialize all $n+1$ tasks within time $T$:  

\begin{quote}
Set local variable $a := a_{frozen} + 1$ {\em (first unused address)};
for all $r \in  \{ r_1, r_2, \ldots, r_{n+1} \}$ set $ip(r) := a$.
{\sc If} {\bf Try ($qp, r_{n+1}, \{ r_1, r_2, \ldots, r_{n+1} \}, 0, \frac{1}{2}$)}
 set $a_{last} := a$ and exit.

  {\em (This means that half the time is assigned
   to all new programs with fresh starts).}
\end{quote}
  
\noindent
{\bf 3.} Set $T := 2 T$, and go to {\bf 1.}
\end{method}
\begin{center}
-----------------------------------------------------------------------------------------
\end{center}

Therefore, given tasks $r_1, r_2, \ldots,$
first initialize $a_{last}$; then
for $i := 1, 2, \ldots $ invoke Realistic {\sc oops}$(i)$  to find
programs starting at (possibly increasing) address $a_{last}$, each
solving all tasks so far, possibly eventually discovering
a universal solver for all tasks in the sequence.

As address $a_{last}$ increases for the $n$-th time,
$q^n$ is defined as
the program starting at $a_{last}$'s old value and
ending right before its new value.
Program $q^m$ ($m>i,j$) may exploit $q^i$ by calling it
as a subprogram, 
or by copying $q^i$ into some state $s(r)$,
then editing it there, e.g., by inserting parts of 
another $q^j$ somewhere,
then executing the edited variant.

\subsection{Near-Bias-Optimality of Realistic OOPS}
\label{near-optimal}

{\sc oops} for realistic computers is not only asymptotically
optimal in the sense of \cite{Levin:73} (see Method \ref{lsearch}),
but also near bias-optimal (compare Def. \ref{bias-optimal},
Observation \ref{fastest}).  To see this,
consider a program $p$ solving the current task set within $k$ steps,
given current code bias $q_{0:a_{frozen}}$ and $a_{last}$.
Denote $p$'s probability by $P(p)$ (compare Eq. (\ref{prob}) and
Method \ref{oopsalg};
for simplicity we omit the obvious conditions).
A bias-optimal solver would find a solution 
within at most $k/P(p)$ steps.
We observe that {\sc oops}
will find a solution within at most $2^3k/P(p)$
steps, ignoring a bit of hardware-specific 
overhead (for marking changed tape components,
measuring time, switching between tasks, etc, compare Section \ref{try}): 
At most a factor 2 might be lost through
allocating half the search time to prolongations of
the most recent code, another factor 2 for the incremental
doubling of $T$ (necessary because we do not know in advance the
best value of $T$), and another factor 2 for {\bf Try}'s
resets of states and tasks.  So the
method is essentially {\em 8-bias-optimal} (ignoring 
hardware issues) with respect to the current task.
If we do {\em not} want to ignore hardware issues:
on currently widely used computers 
we can realistically expect to suffer from slowdown 
factors smaller than acceptable values such as, say, 100.

The advantages of {\sc oops} materialize when 
$P(p) >> P(p')$, where $p'$ is among
the most probable fast solvers of the current task set
that do {\em not} use previously found code.  
Ideally, $p$ is already identical to the most recently frozen code.
Alternatively, 
$p$ may be rather short and thus likely because it uses information
conveyed by  earlier found programs stored below $a_{frozen}$.
For example,
$p$ may call an earlier stored $q^i$  as a subprogram.
Or maybe $p$ is a short and fast program
that copies a large $q^i$ into state $s(r_j)$, then modifies
the copy just a little bit to obtain $\bar{q}^i$, then successfully
applies $\bar{q}^i$ to $r_j$. 
Clearly, if $p'$ is not many times faster than $p$, then
{\sc oops} will in general suffer from a much smaller constant
slowdown factor than {\em non}incremental asymptotically optimal search, 
precisely reflecting the extent to which solutions to successive tasks 
do share useful mutual information,
given the set of primitives for copy-editing them.

Given an optimal problem solver, problem $r$, 
current code bias $q_{0:a_{frozen}}$,
the most recent start address $a_{last}$, 
and information about the starts and ends of previously
frozen programs $q^1,q^2,\ldots,q^k$,
the total search time $T(r,q^1,q^2,\ldots,q^k,a_{last},a_{frozen})$ 
for solving $r$ can be used to define the degree of 
bias 
\[
B(r,q^1,q^2,\ldots,q^k,a_{last},a_{frozen}) :=
1 / T(r,q^1,q^2,\ldots,q^k,a_{last},a_{frozen}).
\]
Compare the concept of {\em conceptual jump size} 
\citep{Solomonoff:86,Solomonoff:89}.

\subsection{Realistic OOPS Variants for Optimization etc.}

Sometimes we are not searching for a universal solver,
but just intend to solve the most recent task $r_{n+1}$.
E.g., for problems of fitness function maximization or
optimization, the $n$-th task
typically is just to find a program than outperforms 
the most recently found program.
In such cases we should use a reduced variant of {\sc oops} which
replaces step {\bf 2} of Method \ref{oopsalg} by:
\begin{quote}
{\bf 2.}
Set $a := a_{frozen} + 1$; set $ip(r_{n+1}) := a$.
{\sc If} {\bf Try ($qp, r_{n+1}, \{ r_{n+1} \}, 0, \frac{1}{2}$)},
 then set $a_{last} := a$ and exit.
\end{quote}
Note that the reduced variant still spends 
significant time on testing earlier solutions:
the probability of any prefix that computes 
the address of some previously frozen program $p$ 
and then calls $p$ determines a lower bound 
on the fraction of total search time spent 
on $p$-like programs.
Compare Observation \ref{startaddress}.

Similar {\sc oops} variants will also assign {\em prewired} 
fractions of the 
total time to the second most recent program and its prolongations,
the third most recent program and its prolongations, etc.
Other {\sc oops} variants will find a program that solves, say, just the 
$m$ most recent tasks, where $m$ is an integer constant, etc.  
Yet other {\sc oops} variants 
will assign more (or less) than half of the 
total time to the most recent code and prolongations thereof.
We may also consider probabilistic {\sc oops} variants 
in {\em Speed-Prior} style 
\citep{Schmidhuber:00v2,Schmidhuber:02colt}.

One not necessarily useful idea: Suppose the number of tasks to be solved by a single
program is known in advance. Now we might think of an {\sc OOPS} variant 
that works on all tasks in parallel, 
again spending half the search time on programs starting at  $a_{last}$,
half on programs starting at  $a_{frozen}+1$;
whenever one of the tasks is solved 
by a prolongation of  $q_{a_{last}:a_{frozen}}$ 
(usually we cannot know in advance which task),
we remove it from the current task ring and 
freeze the code generated so far,
thus increasing $a_{frozen}$
(in contrast to {\bf Try} which
does not freeze programs before the entire current task set is solved). 
If it turns out, however, that not all tasks can be solved
by a program starting at $a_{last}$, we have to 
start from scratch by searching only among programs
starting at $a_{frozen}+1$. Unfortunately, in general we cannot 
guarantee
that this approach of {\em early freezing} will converge.

\subsection{Illustrated Informal Recipe for OOPS Initialization}
\label{recipe}
Given some application, before we can switch 
on {\sc oops} we have to specify our initial bias.

\begin{enumerate}

\item
Given a problem sequence, collect primitives that embody 
the prior knowledge.
Make sure one can interrupt any primitive at any time, and that one
can undo the effects of (partially) executing it.

{\em 
For example, if the task is path planning in a robot simulation,
one of the primitives might be a program that stretches the 
virtual robot's arm until its touch sensors encounter an obstacle. Other 
primitives may include various traditional AI path planners \citep{Russell:94},
artificial neural networks \citep{Werbos:74,Rumelhart:86,Bishop:95}
or support vector machines \citep{Vapnik:92}
for classifying sensory data written into
temporary internal storage, as well as
instructions for repeating the most recent action until some
sensory condition is met, etc.
}

\item
Insert additional prior bias by
defining the rules of an initial probabilistic programming language 
for combining primitives into complex sequential programs.

{\em 
For example, a probabilistic syntax diagram
may specify high probability for executing the robot's
stretch-arm primitive, given some classification 
of a sensory input that was written into temporary, task-specific 
memory by some previously invoked  classifier primitive.
}

\item
To complete the bias initialization,
add primitives for addressing / calling / copying \& editing 
previously frozen programs, and for temporarily modifying
the probabilistic rules of the language (that is, these rules 
should be represented in modifiable task-specific memory as well).
Extend the initial rules of the language to accommodate the
additional primitives.

{\em 
For example, there may be a primitive
that counts the frequency of certain 
primitive combinations in previously frozen programs, 
and temporarily 
increases the probability of the most frequent ones.
Another primitive may conduct a more sophisticated but
also more time-consuming Bayesian analysis,
and write its result into task-specific storage such that
it can be read by subsequent primitives.
Primitives for editing code may invoke variants of 
Evolutionary Computation \citep{Rechenberg:71,Schwefel:74}, 
Genetic Algorithms \citep{Holland:75}, 
Genetic Programming \citep{Cramer:85,Banzhaf:98},
Ant Colony Optimization  \citep{Gambardella:00,Dorigo:99},  etc.
}

\item
Use {\sc oops}, which  invokes {\bf Try}, to bias-optimally
spend your limited computation time on solving 
your problem sequence. 

\end{enumerate}
The experiments (Section \ref{experiments}) 
will use assembler-like primitives that are 
much simpler (and thus in a certain sense less biased) than those 
mentioned in the robot example above. They will suffice, however,
to illustrate the basic principles.

\subsection{Example Initial Programming Language}
\label{seed}

\begin{small}
\hspace{3.9cm}
{\sl ``If it isn't 100 times smaller than 'C' it isn't {\sc Forth}.''} 
 {\sc (Charles Moore)}
\vspace{0.2cm}
\end{small}

The efficient search and backtracking mechanism described
in Section \ref{try} is designed for a broad variety
of possible programming languages, possibly 
list-oriented such as LISP,
or based on matrix operations for recurrent neural network-like
parallel architectures.  Many other alternatives are possible.

A given language
is represented by $Q$, the set of initial tokens. Each token
corresponds to a primitive instruction.
Primitive instructions are computer programs that manipulate tape contents,
to be composed by {\sc oops} such that more complex programs result.
In principle, the ``primitives'' themselves could be large and time-consuming
software, such as, say, traditional AI planners, or theorem provers, 
or multiagent update procedures,
or learning algorithms for neural networks
represented on tapes.  
Compare Section \ref{recipe}.

For each instruction 
there is a unique number between 1 and $n_Q$, such that all such 
numbers are associated with exactly one instruction.
Initial knowledge or bias is 
introduced by writing appropriate primitives and adding them to $Q$. 
Step {\bf 1} of procedure
{\bf Try} (see Section \ref{try}) translates any instruction number
back into the corresponding executable code (in our particular
implementation: a pointer to a $C$-function). If the presently
executed instruction does not directly affect instruction
pointer $ip(r)$, e.g., through a conditional jump, or the
call of a function, or the return from a function call, then
$ip(r)$ is simply incremented.

Given some choice of programming language / initial primitives,
{\bf we typically have to write a new interpreter from scratch,}  instead of 
using an existing one. Why?  Because procedure {\bf Try} (Section \ref{try})
needs total control over 
all (usually hidden and inaccessible)
aspects of storage management, including garbage collection etc. 
Otherwise the storage
clean-up in the wake of executed and tested prefixes could become suboptimal.

For the experiments (Section \ref{experiments})
we wrote (in $C$) an interpreter for an example, stack-based,
universal programming language inspired by {\sc Forth} 
\citep{Forth:70},
whose disciples praise its beauty and the compactness of its programs.

The appendix (Section \ref{language}) 
describes the details.
Data structures on tapes (Section \ref{data})
can be manipulated by primitive instructions listed 
in Sections \ref{stacks}, \ref{control}, \ref{self}.
Section \ref{user} shows how the user may compose complex programs
from primitive ones, and insert them into total code $q$.
Once the user has declared his programs, $n_Q$ will remain fixed.

\section{Limitations and Possible Extensions of OOPS}
\label{miscellaneous}

In what follows we will discuss to which extent
``no free lunch theorems'' are relevant
to {\sc oops} (Section \ref{nfl}),
which are the essential limitations of 
{\sc oops} (Section \ref{limits}),
and how to use {\sc oops} for reinforcement learning
(Section \ref{rl}).

\subsection{How Often Can we Expect to Profit from Earlier Tasks?}
\label{nfl}

How likely is it that any learner can indeed profit from earlier solutions?
At first naive glance this seems unlikely, since it has been well-known for
many decades that
most possible pairs of symbol strings (such as problem-solving programs) 
do not share any algorithmic information; e.g., \cite{LiVitanyi:97}. 
Why not? Most possible combinations
of strings $x,y$ are algorithmically incompressible,
that is, the shortest algorithm computing $y$, given $x$, has
the size of the shortest algorithm computing $y$, given nothing
(typically a bit more than $l(y)$ symbols), {\bf which means 
that $x$ usually does not tell us anything about $y$.} 
Papers in evolutionary computation often mention 
{\em no free lunch theorems} 
\citep{Wolpert:97} which are variations of this ancient
insight of theoretical computer science.

Such at first glance discouraging theorems, however, have a quite 
limited scope: they refer to the very special case of problems sampled 
from i.i.d. uniform distributions on finite problem spaces.  But of course
there are infinitely many distributions besides the uniform one.  
In fact, the uniform one is not only extremely unnatural from
any computational perspective  --- although most objects are random,
computing random objects is much harder than computing nonrandom ones
--- but does not even make sense as we increase data set size and let it
go to infinity: {\em There is no such thing as a uniform distribution
on infinitely many things,} such as the integers.

Typically, successive real world problems are {\bf not}
sampled from uniform distributions.
Instead they tend to be closely related.  In particular,
teachers usually provide sequences of more and more
complex tasks with very similar solutions, and
in optimization the next task typically
is just to outstrip the best approximative solution found so far,
given some basic setup that does not change from one task to the next.
Problem sequences that humans consider to be {\em interesting}
are {\em atypical} when compared to {\em arbitrary} sequences 
of well-defined problems \citep{Schmidhuber:97nn}.
In fact, it is no exaggeration to claim 
that almost the entire field of computer science
is focused on comparatively few atypical
problem sets with exploitable regularities.
For all {\em interesting} problems 
the consideration of previous work is justified,
to the extent that {\em interestingness} implies relatedness
to what's already known \citep{Schmidhuber:02predictable}.
Obviously, {\sc oops}-like procedures are advantageous only where such
relatedness does exist. In any case, however, they will at least 
not do much harm. 

\subsection{Fundamental Limitations of OOPS}
\label{limits}

An appropriate task sequence may help
{\sc oops} to reduce the slowdown factor of
plain {\sc Lsearch} through experience.
Given a single task, however, {\sc oops} does {\bf not} by itself
invent an appropriate series of easier subtasks whose solutions
should be frozen first. 
Of course, since both {\sc Lsearch} and  {\sc oops}
may search in general algorithm space, 
some of the programs they execute 
may be viewed as self-generated subgoal-definers and 
subtask solvers.  But with a single given task
there is no incentive to {\em freeze} intermediate
solutions {\em before} the original task is solved.
The potential speed-up of {\sc oops} {\em does} stem from exploiting
external information encoded within an ordered task sequence. 
This motivates its very name.

Given some final task, a badly chosen training sequence 
of intermediate tasks may cost more search time than
required for solving just the final task by itself, without
any intermediate tasks.

{\sc Oops} is designed for resetable environments.
In {\em non}resetable environments
it loses its theoretical foundation, and becomes a 
heuristic method.  
For example, it is possible to use {\sc oops} for designing optimal
trajectories of robot arms in virtual {\em simulations.}
But once we are working with a real physical robot there may be no 
guarantee that we will be able to precisely reset it 
as required by backtracking procedure {\bf Try}.

{\sc Oops} neglects one source of potential speed-up relevant for
reinforcement learning \citep{Kaelbling:96}:
it does not predict future tasks from previous ones,
and does not spend a fraction of its time on 
solving predicted tasks.  Such issues will be
addressed in the next subsection.

\subsection{Outline of OOPS-based Reinforcement Learning (OOPS-RL)}
\label{rl}

At any given time, a reinforcement learner \citep{Kaelbling:96} will
try to find a {\em policy} (a strategy for future decision
making) that maximizes its expected future reward.
In many traditional reinforcement learning (RL) applications,
the policy that works best in a given set of training trials
will also be optimal in future test trials \citep{Schmidhuber:01direct}.
Sometimes, however, it won't.  To see the difference between
searching (the topic of the previous sections) and
reinforcement learning (RL), consider an agent and two boxes.
In the $n$-th trial the agent may open and collect the content of
exactly one box. The left box will contain
$100n$ Swiss Francs, the right box $2^n$ Swiss Francs,
but the agent does not know this
in advance. During the first 9 trials
the optimal policy is {\em ``open left box.''} This is what a good
searcher should find, given the outcomes of the first 9 trials.
But this policy will be suboptimal in trial 10.
A good reinforcement learner, however, should extract the
underlying regularity in the reward generation process
and predict the future reward, picking the right box in
trial 10, without having seen it yet.
 
 The first general, asymptotically optimal reinforcement learner is
 the recent AIXI model \citep{Hutter:01aixi,Hutter:02selfopt}.
 It is valid for a very broad class of
 environments whose reactions to action sequences (control signals)
 are sampled from arbitrary computable probability distributions.
 This means that AIXI is far more general than traditional RL approaches.
 However, while AIXI clarifies the theoretical limits of
 RL, it is not practically feasible, just like {\sc Hsearch} is not.
 From a pragmatic point of view,
 what we are really interested in is a reinforcement learner
 that makes optimal use of given, limited computational resources.
 In what follows, we will outline how to use {\sc oops}-like bias-optimal
 methods as components of universal yet feasible reinforcement learners.
  
  We need two {\sc oops} modules. The first is
  called the predictor or world model. The second is
  an action searcher using the world model.  The life of the entire system
  should consist of a sequence of {\em cycles} 1, 2, ...
  At each cycle, a limited amount of computation time will
  be available to each module.
  For simplicity we assume that during each cyle the system
  may take exactly one action.
  Generalizations to actions consuming 
  several cycles are straight-forward though.
  At any given cycle, the system executes the following procedure:

\begin{enumerate}
\item
For a time interval fixed in advance,
the predictor is first trained
in bias-optimal fashion to find a better world model,
that is, a program that predicts the inputs from the environment
(including the rewards, if there are any), given a history of
previous observations and actions.
So the $n$-th task ($n=1,2,\ldots$)
of the first {\sc oops} module is to find (if possible) a
better predictor than the best found so far.
 
\item
Once the current cycle's time for predictor improvement is used up,
the current world model (prediction program) found by
the first {\sc oops} module will be used by
the second module, again in bias-optimal fashion,
to search for a future action sequence that maximizes
the predicted cumulative reward (up to some time limit).
That is, the $n$-th task ($n=1,2,\ldots$)
of the second {\sc oops} module will be to find
a control program that computes a control sequence of actions,
to be fed into the program representing the current world model
(whose input predictions are successively fed back to itself in
the obvious manner),
such that this control sequence leads to higher predicted
reward than the one generated by the best control program
found so far.
  
\item
Once the current cycle's time for control program search
is used up, we will execute the current action of the best control
program found in step 2. Now we are ready for the next cycle.
   
\end{enumerate}
The approach is reminiscent of an earlier, heuristic, non-bias-optimal
RL approach based on two adaptive recurrent neural networks,
one representing the world model, the other one a controller
that uses the world model to extract a policy for maximizing
expected reward \citep{Schmidhuber:91nips}. The method was inspired by
previous combinations of {\em non}recurrent,
{\em reactive} world models and controllers
\citep{Werbos:87specifications,NguyenWidrow:89,JordanRumelhart:90}.

At any  given time, until which temporal horizon should the predictor
try to predict?  In the AIXI case, the proper way of treating the
temporal horizon
is not to discount it exponentially, as done in most traditional
work on reinforcement learning, but to let the future horizon grow
in proportion to the learner's lifetime so far \citep{Hutter:02selfopt}.
It remains to be seen whether this insight carries over to {\sc oops}-based RL.
In particular, is it possible to prove that variants of
OOPS-RL as above are a near-bias-optimal way of spending a given amount of
computation time on RL problems? Or should we instead combine
{\sc oops} and Hutter's time-bounded AIXI$(t,l)$ model?  
We observe that certain important problems are still open.

\section{Experiments}
\label{experiments}

Experiments can tell us something about the usefulness
of a particular initial bias such as the one 
incorporated by a particular
programming language with particular initial instructions.
In what follows we will describe illustrative problems and 
results obtained using the {\sc Forth}-inspired language
specified in the appendix (Section \ref{language}). The latter
should be consulted for 
the details of the instructions appearing in 
programs found by {\sc oops}. 

While explaining the learning system's setup,
we will also try to identify several
more or less hidden sources of initial bias.

\subsection{On Task-Specific Initialization}
\label{init}

Besides the $61$ initial primitive instructions from Sections
\ref{stacks}, \ref{control}, \ref{self} (appendix),
the only user-defined (complex) tokens are those declared in
Section \ref{user} 
(except for the last one, {\sc tailrec}). That is, we have a total of $61+7=68$
initial non-task-specific primitives.

Given any task, we add task-specific instructions.
In the present experiments, we do {\em not} provide a {\em probabilistic
syntax diagram} defining 
conditional probabilities of certain tokens, given previous tokens. 
Instead we simply start with a maximum entropy distribution on the 
$n_Q >68$ tokens $Q_i$, initializing all probabilities
$p_i= \frac{1}{n_Q}$,
setting all $p[curp][i]:=1$ and $sum[curp]:=n_Q$ 
(compare Section \ref{data}). 

Note that the instruction numbers themselves significantly affect
the initial bias. Some instruction numbers, in particular the small 
ones, are computable by very short programs, others are not.
In general, programs consisting of many instructions that are not
so easily computable, given the initial arithmetic instructions (Section \ref{stacks}),
tend to be less probable. Similarly, as the number of frozen
programs grows, those with higher addresses in general
become harder to access, that is, the address computation
may require longer subprograms.

For the experiments we {\em insert substantial prior bias} by
assigning the lowest (easily computable) instruction numbers to
the task-specific instructions, and by {\bf boosting} (see instruction
{\em boostq} in Section \ref{self}) the 
appropriate {\em ``small number pushers''}
(such as {\em c1, c2, c3}; compare Section \ref{stacks}) 
that push onto data stack {\em ds}
the numbers of the task-specific instructions, such that they
become executable as part of code on {\em ds}.
We also boost the simple arithmetic instructions
{\em by2} (multiply top stack element by 2) 
and {\em dec} (decrement top stack element), 
such that the system can easily
create other integers from the probable ones. For example,
without these boosts the 
code sequence {\em (c3 by2 by2 dec)} (which returns integer 11)
would be much less likely.  Finally we 
express our initial belief in the occasional
future usefulness of previously useful instructions,
by also boosting {\em boostq} itself. 

The following numbers represent maximal values 
enforced in the experiments: 
state size: $l(s)=3000$;
absolute tape cell contents $s_i(r)$: $10^9$;   
number of self-made functions: $100$,
of self-made search patterns or probability distributions per tape: $20$;
callstack pointer: $maxcp=100$;
data stack pointers: $maxdp=maxDp=200$.

\subsection{Towers of Hanoi: the Problem}
\label{hanoi}

Given are $n$ disks of $n$ different sizes,
stacked in decreasing size on the first of three pegs. 
One may move some peg's top disk to the top of another peg, 
one disk at a time, but never a larger disk onto a smaller. The goal
is to transfer all disks to the third peg. Remarkably, 
the fastest way of solving this famous problem requires $2^n - 1$ moves
$(n \geq 0)$.  

The problem is of the {\em reward-only-at-goal} type --- 
given some instance of size $n$, there is no intermediate reward for
achieving instance-specific subgoals.

The exponential growth of minimal solution size is what
makes the problem interesting: Brute force methods searching
in raw solution space will quickly fail as $n$ increases.
But the rapidly growing
solutions do have something in common, namely, the short algorithm
that generates them. Smart searchers will exploit such
algorithmic regularities. 
Once we are searching in general algorithm space,
however, it is essential to efficiently allocate time 
to algorithm tests. This is what {\sc oops} does, in
near-bias-optimal incremental fashion.

Untrained humans find it hard to solve instances $n>6$.
\cite{Anderson:86}
applied traditional reinforcement learning methods 
and was able to solve instances up to $n=3$, solvable within
at most 7 moves.  \cite{Langley:85} used 
learning production systems and was able to solve instances up to $n=5$, 
solvable within at most 31 moves.  
({\em Side note:} 
\cite{Baum:99} also applied an alternative reinforcement
learner based on the artificial economy by \cite{Holland:85}
to a simpler 3 peg blocks world problem
where any disk may be placed on any other;
thus the required number of moves grows only 
linearly with the number of disks, not exponentially; 
\cite{Kwee:01market} were able to 
replicate their results for $n$ up to 5.)
Traditional AI planning procedures---compare
chapter V by \cite{Russell:94} and \cite{Jana:97}---do not 
learn but systematically explore all possible move combinations,
using only absolutely necessary task-specific primitives 
(while {\sc oops} will later use more than 70 general instructions, most of
them unnecessary).  On current personal computers
AI planners tend to fail to solve Hanoi problem instances with $n > 15$ 
due to the exploding search space (Jana Koehler, IBM Research, 
personal communication, 2002).
{\sc oops}, however, searches program space instead of
raw solution space. Therefore, in principle it should be able to solve
arbitrary instances by discovering
the problem's elegant recursive solution---given $n$ and 
three pegs $S,A,D$ (source peg, auxiliary peg, destination peg), 
define the following procedure:

\begin{quote} 
{\em 
{\sc hanoi}(S,A,D,n): {\sc If} $n=0$ exit;  {\sc Else Do}: 

call {\sc hanoi}(S, D, A, n-1); 
move top disk from S to D; 
call {\sc hanoi}(A, S, D, n-1).
}
\end{quote}

\subsection{Task Representation and Domain-Specific Primitives}
\label{setup}

The $n$-th problem is to solve all Hanoi instances up
to instance $n$.  Following our general rule, we represent the 
dynamic environment for task $n$ on the $n$-th task tape, 
allocating $n+1$ addresses for each peg, 
to store the order and the sizes of its current disks, and 
a pointer to its top disk (0 if there isn't one).

We represent pegs $S, A, D$ by numbers 1, 2, 3, respectively.
That is, given an instance of size $n$, we push onto data stack {\em ds} the values
$1, 2, 3, n$. 
By doing so we insert {\em substantial, nontrivial
prior knowledge} about the fact
that it is useful to represent each peg by a symbol, and to
know the problem size in advance. The task is completely 
defined by $n$; the other 3 values are just useful for the following
primitive instructions added to
the programming language of Section \ref{language}:
Instruction {\em mvdsk()} assumes that 
$S, A, D$ are represented by the first three elements  
on data stack {\em ds} above the current base pointer $cs[cp].base$  (Section \ref{data}).
It operates in the obvious fashion
by moving a disk from peg $S$ to peg $D$.
Instruction {\em xSA()} exchanges the representations of $S$ and $A$,
{\em xAD()} those of $A$ and $D$ (combinations may create arbitrary 
peg patterns).
Illegal moves cause the current program prefix to halt. 
Overall success is easily verifiable since
our objective is achieved once the first two pegs are empty.

\subsection{Incremental Learning: First Solve Simpler Context Free Language Tasks}
\label{1n2n}

Despite the near-bias-optimality of {\sc oops}, 
within reasonable time (a week) on a
personal computer, the system with 71 initial instructions 
was not able to solve
instances involving more than 3 disks. 
What does this mean? {\em Since search time of an optimal searcher is
a natural measure of initial bias,} it just means that the already 
nonnegligible bias towards our task set was still too weak.

This actually gives us an opportunity to demonstrate that  {\sc oops}
can indeed benefit from its incremental learning abilities.
Unlike Levin's and Hutter's {\em non}incremental methods it 
always tries to profit from experience with previous tasks.
Therefore, to properly illustrate its behavior, we need an example where it {\em does} profit.
In what follows, we will first train it on additional, easier
tasks, to teach it something about recursion, hoping that the resulting
code bias shifts will help to solve the Hanoi tasks as well.

For this purpose we use a seemingly unrelated problem class based on
the context free language $\{1^n2^n\}$: given input
$n$ on the data stack {\em ds}, the goal is to
place symbols on the auxiliary stack {\em Ds} such that 
the $2n$ topmost elements are $n$ 1's followed by $n$ 2's.
Again there is no intermediate reward 
for achieving instance-specific subgoals.

After every executed instruction we test whether the objective has been achieved.
By definition, the time cost per test (measured in unit time steps; Section \ref{primitives})
equals the number of considered elements of {\em Ds}.
Here we have an example
of a test that may become more expensive with instance size.

We add two more instructions to the initial programming language:
instruction {\em 1toD()} pushes 1 onto {\em Ds}, instruction {\em 2toD()} pushes 2.
Now we have a total of five task-specific instructions (including those
for Hanoi), with instruction numbers 1, 2, 3, 4, 5, for
{\em 1toD}, {\em 2toD}, {\em mvdsk}, {\em xSA}, {\em xAD}, respectively,
which gives a total of 73 initial instructions.

So we first boost
(Section \ref{self})
the ``small number pushers'' {\em c1, c2} (Section \ref{stacks}) 
for the first training phase where
the $n$-th task $(n=1, \ldots, 30)$ is to solve all $1^n2^n$ problem
instances up to $n$.
Then we undo the $1^n2^n$-specific boosts of {\em c1, c2}
and boost instead
 the Hanoi-specific instruction number pushers $c3, c4, c5$
  for the subsequent training phase where
   the $n$-th task (again $n=1, \ldots, 30$)
    is to solve all Hanoi instances up to $n$.

\subsection{C-Code}
All of the above 
was implemented by a dozen pages of 
code written in C, mostly comments and documentation:
Multitasking and storage management 
through an iterative variant of {\em round robin} {\bf Try} (Section \ref{try}); 
interpreter and 62 basic instructions (Section \ref{language}); 
simple user interface for complex declarations (Section \ref{user}); 
applications to $1^n2^n$-problems (Section \ref{1n2n})
and Hanoi problems (Section \ref{hanoi}).
The current nonoptimized
implementation considers between one and two million discrete
unit time steps per second on an off-the-shelf PC (1.5 GHz).

\subsection{Experimental Results for Both Task Sets}

Within roughly 0.3 days, {\sc oops} found and
froze code solving all thirty $1^n2^n$-tasks.
Thereafter, within 2-3 additional days, it also found a
universal Hanoi solver. The latter does not call the $1^n2^n$ 
solver as a subprogram (which would not make sense at all), but it
does profit from experience: it
begins with a rather short prefix that reshapes the distribution 
on the possible suffixes, an thus the search space,
by temporally increasing the probabilities of
certain instructions of the  earlier found $1^n2^n$ solver. 
This in turn happens to increase the probability of finding a Hanoi-solving suffix.
It is instructive to study the sequence of intermediate solutions.
In what follows we will transform integer sequences discovered
by {\sc oops} back into readable programs
(compare instruction details in Section \ref{language}).

\begin{enumerate}

\item
For the $1^n2^n$-problem, 
within 480,000 time steps (less than a second), {\sc oops} found 
nongeneral but working code for $n=1$: {\em (defnp 2toD).}

\item
At time $10^7$ (roughly 10 {\em s}) it had solved the 2nd instance
by simply prolonging the previous code, using the old, unchanged
start address $a_{last}$: {\em (defnp 2toD grt c2 c2 endnp).}
So this code solves the first two instances.

\item
At time $10^8$ (roughly 1 {\em min}) it had solved the 3rd instance, again
through prolongation:

{\em (defnp 2toD grt c2 c2 endnp  boostq delD delD bsf 2toD).}

Here instruction {\em boostq} greatly boosted 
the probabilities of the subsequent instructions.

\item
At time $2.85*10^9$ 
(less than 1 hour)
it had solved the 4th instance through prolongation:

{\em (defnp 2toD grt c2 c2 endnp  boostq delD delD bsf 2toD fromD delD delD delD fromD 
bsf by2 bsf).}

\item
At time $3*10^9$ (a few minutes later)
it had solved the 5th instance through prolongation:

{\em (defnp 2toD grt c2 c2 endnp  boostq delD delD bsf 2toD fromD delD delD delD fromD 
bsf by2 bsf  by2 fromD delD delD fromD cpnb bsf).}

The code found so far was lengthy and unelegant. But it does solve the first
5 instances.

\item
Finally, at time $30,665,044,953$ (roughly 0.3 days), {\sc oops} had
      created and tested a new, elegant, recursive
           program (no prolongation of the previous one) 
	   with a new increased start address
	        $a_{last}$, solving all instances up to 6:
		     {\em (defnp c1 calltp c2 endnp).}

That is, it was cheaper to solve all instances up to 6 by discovering
and applying this new program to all instances so far, than just prolonging 
the old code on instance 6 only.

\item
The program above turns out to be a near-optimal universal $1^n2^n$-problem solver.
On the stack, it
constructs a 1-argument procedure that returns nothing
if its input argument is 0,
otherwise calls the instruction {\em 1toD} whose code is 1, then calls
itself with a decremented input argument, then calls {\em 2toD} whose code is 2,
then returns.  

That is, all remaining  $1^n2^n$-tasks can profit from the solver of instance 6.
Reusing this current program $q_{a_{last}:a_{frozen}}$ again and again,
within very few additional time steps (roughly 20 milliseconds), 
by time $30,665,064,543$,
{\sc oops} had also solved the remaining 24 $1^n2^n$-tasks up to $n=30$.

\item
Then {\sc oops} switched to the Hanoi problem.
Almost immediately (less than 1 {\em ms} later), at time $30,665,064,995$,
it had found the trivial code for $n=1$:
{\em (mvdsk).}

\item
Much later, by time $260 *10^9$ (more than 1 day),
it had found fresh yet somewhat bizarre
code (new start address $a_{last}$) for $n=1,2$:
{\em (c4 c3 cpn c4 by2 c3 by2 exec).}

The long search time so far indicates that the 
Hanoi-specific bias still is not very high.

\item
Finally, by time $541 *10^9$ (roughly 3 days), 
it had found fresh code (new $a_{last}$) for $n=1,2,3$:

{\em (c3 dec boostq defnp c4 calltp c3 c5 calltp endnp).}

\item
The latter turns out to be a near-optimal
universal Hanoi solver, and greatly profits from the  code bias embodied
by the earlier found $1^n2^n$-solver (see analysis in Section \ref{discussion} below). 
Therefore, by time $679*10^9$,
{\sc oops} had solved the remaining 27 tasks for $n$ up to 30, reusing  
the same  program $q_{a_{last}:a_{frozen}}$ again and again.

\end{enumerate}
The entire 4-day search for solutions to all 60 tasks tested
93,994,568,009 
prefixes corresponding to
345,450,362,522
instructions
costing 
678,634,413,962 
time steps.
{\em Recall once more that
search time of an optimal solver is a natural measure
of initial bias.}
Clearly, most tested prefixes are short --- they either 
halt or get interrupted soon.
Still, some programs do run for a long time; for example, 
the run of the self-discovered universal Hanoi solver working 
on instance 30 
consumed 33 billion steps, 
which is already 5 \% of the total time.
The stack used by the iterative equivalent of
procedure {\bf Try} for storage management (Section \ref{try})
never held more than 20,000 elements though.

\subsection{Analysis of the Results}
\label{discussion}


The final 10-token Hanoi solution demonstrates the benefits of
incremental learning: it greatly profits from
rewriting the search procedure with the help of information 
conveyed by the earlier recursive solution to the $1^n2^n$-problem. How?

The prefix {\em (c3 dec boostq)} (probability 0.003)
prepares the foundations: 
Instruction {\em c3} pushes 3; {\em dec}
decrements this; {\em boostq} takes the result 2 as an argument
(interpreted as an address)
and thus boosts the probabilities of all components of the 2nd
frozen program, which happens to be the previously found universal
$1^n2^n$-solver.  This causes an online bias shift 
on the space of possible suffixes:
it greatly increases the probability that
{\em defnp, calltp, endnp,} will appear
in the remainder of the online-generated program.
These instructions in turn are helpful for building
(on the data stack {\em ds})
the double-recursive procedure
generated by the suffix
{\em (defnp c4 calltp c3 c5 calltp endnp),} which essentially
constructs (on data stack $ds$) 
a 4-argument procedure that returns nothing if its input argument is 0,
otherwise decrements the top input argument, calls the instruction
{\em xAD} whose code is 4,
then calls itself on a copy of the top 4 arguments,
then calls {\em mvdsk} whose code is 5,
then calls {\em xSA} whose code is 3,
then calls itself on another copy of the top 4 arguments,
then makes yet another (unnecessary) argument copy, then returns
(compare the standard Hanoi solution).

The total probability of the final solution, given the previous
codes, is calculated as follows: 
since $n_Q=73$, given the boosts of
{\em c3, c4, c5, by2, dec, boostq},
we have probability 
$( \frac{1+73}{7 * 73})^3$ for the prefix {\em (c3 dec boostq)};
since this prefix further boosts {\em defnp, c1, calltp, c2, endnp,}
we have probability 
$( \frac{1+73}{12 * 73})^7$ for the suffix {\em (defnp c4 calltp c3 c5 calltp endnp)}.
That is, the probability of the complete 10-symbol code is $9.3 * 10^{-11}$. 
On the other hand, the probability of the essential Hanoi-specific suffix
{\em (defnp c4 calltp c3 c5 calltp endnp)}, given just the initial boosts,
is only $( \frac{1+73}{7 * 73})^3 ( \frac{1}{7 * 73})^4 =   4.5 * 10^{-14}$, 
which explains why it was not
quickly found without the help of the solution to an easier problem set. 
(Without any initial boosts its probability would actually have 
been similar: $(\frac{1}{73})^7=9 * 10^{-14}$.)
This would correspond to a search time of several years, 
as opposed to a few days. 

So in this particular setup the simple recursion for the $1^n2^n$-problem
indeed provided useful incremental training for the more complex Hanoi recursion,
speeding up the search by a factor of 1000 or so.

On the other hand, the search for the universal solver for all $1^n2^n$-problems 
(first found with instance $n=6$) 
did not at all profit from solutions to earlier solved tasks 
(although instances $n>6$ did profit).

Note that the time spent by the final 10-token Hanoi solver
on increasing the probabilities of certain instructions and
on constructing executable code on the data stack (less than 
50 time steps) quickly becomes negligible as 
the Hanoi instance size grows. In this particular application,
most time is spent on
executing the code, not on constructing it.

Once the universal Hanoi solver was discovered, why
did the solution of the remaining Hanoi tasks
substantially increase the total time (by roughly 25 \%)? 
Because the sheer {\em runtime} of the discovered, frozen, 
near-optimal program on the remaining tasks was already comparable to 
the previously consumed {\em search time} for this program, due to the very
nature of the Hanoi task:
Recall that a solution for $n=30$ takes more than a billion {\em mvdsk} 
operations, and that for each {\em mvdsk} several other instructions need to be
executed as well. Note that experiments with traditional reinforcement 
learners \citep{Kaelbling:96} rarely involve problems whose 
solution sizes exceed a few thousand steps.

Note also that we could continue to solve Hanoi tasks up to $n>40$.
The execution time required to solve such instances with an optimal solver
greatly exceeds the search time required for finding the solver itself.
There it does not matter much whether {\sc oops}
already starts with a prewired Hanoi solver, or
first has to discover one, since the initial search time for the
solver becomes negligible anyway.

Of course, different initial bias can yield dramatically different results.
For example, using hindsight we could set to zero the probabilities
of all 73 initial instructions  (most are unnecessary for the 30
Hanoi tasks) except for the 7 instructions used 
by the Hanoi-solving suffix, then make those 7 instructions equally likely, 
and thus obtain a comparatively high Hanoi solver
probability of $(\frac{1}{7})^7 =   1.2 * 10^{-6}$. 
This would allow for finding the solution to the 10 disk Hanoi problem
within less than an hour, without having to learn easier tasks first  
(at the expense of obtaining a nonuniversal
initial programming language).
The point of this experimental section, however, is {\bf not}
to find the most
reasonable initial bias for particular problems, but to illustrate
the basic functionality of the first general, near-bias-optimal,  
incremental learner.

{\bf Future research} 
may focus on devising particularly compact,
particularly reasonable sets of initial codes with
particularly broad practical applicability. 
It may turn out that the most useful initial
languages are not traditional programming 
languages similar to the {\sc Forth}-like
one from Section \ref{language}, but instead
based on a handful of primitive instructions for
massively parallel cellular automata 
\citep{Ulam:50,Neumann:66,Zuse:69,Wolfram:84}, 
or on a few nonlinear operations on matrix-like
data structures such as those used in
recurrent neural network research 
\citep{Werbos:74,Rumelhart:86,Bishop:95}.
For example, we could use the principles of
{\sc oops} to create a non-gradient-based,  near-bias-optimal
variant of the successful recurrent network 
metalearner by  \cite{Hochreiter:01meta}.
It should also be of interest to study probabilistic 
{\em Speed Prior}-based {\sc oops} variants \citep{Schmidhuber:02colt}
and to devise applications of {\sc oops}-like methods as
components of universal reinforcement learners  (Section \ref{rl}).
In ongoing work, we are applying {\sc oops} to the problem
of  optimal trajectory planning for robotics
in a realistic physics simulation. 
This involves the interesting trade-off
between comparatively fast program-composing primitives or
{\em ``thinking primitives''} 
and time-consuming {\em ``action primitives''},
such as {\em stretch-arm-until-touch-sensor-input}
(compare Section \ref{recipe}).

\subsection{Physical Limitations of OOPS}
\label{outlook}

Due to its generality and its optimality properties, 
{\sc oops} should scale to large problems in
an essentially unbeatable fashion, thus raising the question: 
Which are its physical limitations?
To give a very preliminary answer, we first observe
that with each decade computers become roughly 1000 times faster by cost,
reflecting Moore's empirical law first formulated in  1965.
Within a few decades {\em non}reversible computation will encounter
fundamental heating problems associated with high density computing
\citep{Bennett:82}. Remarkably, however, {\sc oops} 
can be naturally implemented using the {\em reversible} computing 
strategies \citep{Fredkin:82}, since it completely resets all state 
modifications due to the programs it tests.
But even when we naively extrapolate Moore's law, 
within the next century {\sc oops} will hit
the limit of \cite{Bremermann:82}:
approximately $10^{51}$ operations per second on 
$10^{32}$ bits for the ``ultimate laptop'' \citep{Lloyd:00} with 1 kg of
mass and 1 liter of volume. Clearly, the Bremermann limit constrains 
the maximal {\em conceptual jump size}
\citep{Solomonoff:86,Solomonoff:89} from one problem to the next.
For example, given some prior code bias derived from solutions
to previous problems,
within 1 minute, a sun-sized {\sc oops} (roughly $2 \times 10^{30} kg$)
might be able to solve an additional problem that requires finding 
an additional 200 bit program with, say, $10^{20}$ steps runtime.  
But within the next centuries, 
{\sc oops} will fail on new problems that require 
additional 300 bit programs of this type, 
since the speed of light greatly limits the acquisition 
of additional mass, through a function quadratic in time.

Still, even the comparatively modest hardware speed-up 
factor $10^9$ expected for the next 30 years appears quite promising 
for {\sc oops}-like systems. For example,
with the 73 token language used in the experiments (Section \ref{experiments}),
we could learn from scratch (within a day or so) 
to solve the 20 disk Hanoi problem ($>10^6$ moves), 
without any need for 
boosting task-specific instructions, or for incremental search through instances $< 20$, 
or for additional training sequences of easier tasks.  
Comparable speed-ups will be achievable much earlier by 
distributing {\sc oops} across large computer networks 
or by using supercomputers---on the fastest current machines 
our 60 tasks (Section \ref{experiments})
should be solvable within a few seconds as 
opposed to 4 days.

\subsection*{Acknowledgments}
Thanks to 
Ray Solomonoff,
Marcus Hutter,
Sepp Hochreiter,
Bram Bakker, 
Alex Graves, 
Douglas Eck, 
Viktor Zhumatiy, 
Giovanni Pettinaro,
Andrea Rizzoli,
Monaldo Mastrolilli,
Ivo Kwee,
and several unknown NIPS referees,
for useful discussions or
helpful comments on drafts or short versions of
this paper,
to Jana Koehler for sharing her insights
concerning AI planning procedures,
and to Philip J. Koopman Jr. for granting permission to
reprint the quote in Section \ref{control}.
Hutter's frequently mentioned work was funded through the
author's SNF grant 2000-061847
``Unification of universal inductive inference and
sequential decision theory.''

\appendix 
\section{Example Programming Language}
\label{language}

{\sc Oops} can be seeded with a wide variety of programming languages.
For the experiments, we wrote an interpreter for a 
stack-based universal programming language inspired
by {\sc Forth} \citep{Forth:70}.
We provide initial instructions for
defining and calling recursive functions, iterative loops, arithmetic
operations, and domain-specific behavior.  
Optimal metasearching for better search algorithms is
enabled through bias-shifting instructions that can modify
the conditional probabilities of future search options in currently
running self-delimiting programs. 
Sections \ref{data},
explains the basic data structures;
Sections \ref{stacks},
\ref{control},
\ref{self}
define basic primitive instructions;
Section \ref{user} shows how to compose complex programs
from primitive ones, and explains how the user may insert them into total code $q$.

\subsection{Data Structures on Tapes}
\label{data}
\begin{table*}

\label{specific}
\begin{center}
\begin{tabular}{|c|c|c|c|c|c|c|}    \hline
{\bf Symbol} & {\bf Description} \\  \hline
{\em ds} & data stack holding arguments of functions, possibly also edited code \\  \hline
{\em dp} & stack pointer of {\em ds} \\  \hline
{\em Ds} & auxiliary data stack \\  \hline
{\em Dp} & stack pointer of {\em Ds} \\  \hline
{\em cs} & call stack or runtime stack to handle function calls \\  \hline
{\em cp} & stack pointer of {\em cs} \\  \hline
$cs[cp].ip$  & current function call's instruction pointer $ip(r) := cs[cp](r).ip$ \\  \hline
$cs[cp].base$  & current base pointer into {\em ds} right below the current input arguments \\  \hline
$cs[cp].out$  & number of return values expected on top of {\em ds} above $cs[cp].base$ \\  \hline
{\em fns} & stack of currently available self-made functions \\  \hline
{\em fnp} & stack pointer of {\em fns} \\  \hline
$fns[fnp].code$  & start address of code of most recent self-made function \\  \hline
$fns[fnp].in$  & number of input arguments of most recent self-made function \\  \hline
$fns[fnp].out$  & number of return values of most recent self-made function \\  \hline
{\em pats} & stack of search patterns (probability distributions on $Q$) \\  \hline
{\em patp} & stack pointer of {\em pats} \\  \hline
{\em curp} & pointer to current search pattern in {\em pats}, $0 \leq curp \leq patp$  \\  \hline
$p[curp][i]$  & $i$-th numerator of current search pattern \\  \hline
$sum[curp]$  & denominator; the current probability of $Q_i$ is $p[curp][i] / sum[curp]$  \\  \hline
\end{tabular}
\end{center}
\caption{{\em
Frequently used implementation-specific symbols,
relating to the data structures used by 
a particular {\sc Forth}-inspired programming language  (Section \ref{language}).
{\bf Not} necessary for understanding the basic principles of {\sc oops}.
}}
\end{table*}

Each tape $r$ contains various stack-like data structures represented
as sequences of integers.
For any stack $Xs(r)$ introduced below 
(here $X$ stands for a character string reminiscent of the stack type)
there is a (frequently not even mentioned) stack 
pointer $Xp(r)$; $0 \leq Xp(r) \leq maxXp$,
located at address $a_{Xp}$, and initialized by 0.
The $n$-th element of  $Xs(r)$ is denoted $Xs[n](r)$.
For simplicity we will often omit tape indices $r$.
Each tape has:

\begin{enumerate}

\item
A data stack {\em ds(r)} (or {\em ds} for short, omitting the task index)
for storing function arguments. (The corresponding
stack pointer  is $dp: 0 \leq dp \leq maxdp$).

\item
An auxiliary data stack {\em Ds}.

\item
A runtime stack or {\em callstack} $cs$ for handling (possibly recursive) functions.
Callstack pointer $cp$ is initialized by 0 for the ``main'' program.
The $k$-th callstack entry ($k = 0, \ldots, cp$) contains 3 variables:
an instruction pointer $cs[k](r).ip$ (or simply $cs[k].ip$, omitting task index $r$) 
initialized by the 
start address of the code of some procedure $f$,
a pointer $cs[k].base$ pointing into {\em ds} right below the values
which are considered input arguments of $f$, 
and the number $cs[k].out$ of return values 
$ds[cs[k].base+1], \ldots, ds[dp]$ 
expected on top of {\em ds} once $f$ has returned. 
$cs[cp]$ refers to the topmost entry
containing the current instruction pointer $ip(r):=cs[cp](r).ip$.

\item
A stack {\em fns} of entries describing self-made functions.
The entry for function {\em fn} contains 
3 integer variables: the start address of {\em fn}'s code,
the number of input arguments expected by {\em fn} on top of {\em ds},
and the number of output values to be returned.

\item
A stack {\em pats} of search patterns.  $pats[i]$ 
stands for a probability distribution on 
search options (next instruction candidates).
It is represented by 
$n_Q+1$ integers 
$p[i][n]$ ($1 \leq n \leq n_Q$) 
and 
{\em sum[i]} (for efficiency reasons).
Once $ip(r)$ hits the current search address $l(q)+1$,
the history-dependent probability of the $n$-th possible
next instruction $Q_n$ 
(a candidate value for $q_{ip(r)}$) 
is given by $p[curp][n] / sum[curp]$,
where $curp$ is another tape-represented variable
($0 \leq curp \leq patp$) indicating the current search pattern.

\item
A binary {\em quoteflag} 
determining whether the instructions pointed to by {\em ip}
will get executed or just {\em quoted}, that is, pushed onto {\em ds}.

\item
A variable holding the index $r$ of this tape's task.

\item
A stack of integer arrays, each having a name, an address, and a size (not used
in this paper, but implemented and mentioned for the sake of completeness).

\item
Additional problem-specific dynamic data structures for 
problem-specific data, e.g., to represent changes of the environment.
An example environment for the {\em Towers of Hanoi}
problem is described in Section \ref{experiments}.

\end{enumerate}

\subsection{Primitive Instructions}
\label{primitives}

Most of the 61 tokens below do not appear in the solutions
found by {\sc oops} in the experiments (Section \ref{experiments}).
Still, we list all of them for completeness' sake, and to provide
at least one example way of seeding {\sc oops} with an initial
set of behaviors.

In the following subsections,
any instruction of the form {\em inst ($x_1, \ldots, x_n$)}
expects its $n$ arguments 
on top of data stack {\em ds}, and replaces
them by its return values, adjusting {\em dp} accordingly
--- the form
{\em inst()} is used for instructions without arguments. 

Illegal use of any instruction will cause the currently considered 
program prefix to halt. In particular, it is illegal 
to set variables (such as stack pointers or instruction pointers)
to values outside their prewired given ranges,
or to pop empty stacks, or to divide by zero, 
or to call a nonexistent function, etc.

Since CPU time measurements on our PCs turned out to be unreliable,
we defined our own, rather realistic time scales.
By definition, most instructions listed below cost exactly 
1 unit time step.  Some, however, consume more time: Instructions
making copies of strings with length $n$ 
(such as {\em cpn(n)}) 
cost $n$ time steps; 
so do instructions 
(such as {\em find(x)})
accessing an {\em a priori} unknown number $n$ of tape cells; 
so do instructions (such as {\em boostq(k)})
modifying the probabilities of 
an {\em a priori} unknown number $n$ of instructions.

\subsubsection{Basic Data Stack-Related Instructions}
\label{stacks}

\begin{enumerate}

\item {\sc Arithmetic.}
{\em c0(),c1(), c2(), ..., c5()} return constants 0, 1, 2, 3, 4, 5, respectively;
{\em inc(x)} returns $x+1$;
{\em dec(x)} returns $x-1$;
{\em by2(x)} returns $2x$;
{\em add(x,y)} returns $x+y$;
{\em sub(x,y)} returns $x-y$;
{\em mul(x,y)} returns $x*y$;
{\em div(x,y)} returns the smallest integer $\leq x/y$;
{\em powr(x,y)} returns $x^y$ (and costs $y$ unit time steps).

\item {\sc Boolean.}
Operand {\em eq(x,y)} returns 1 if $x=y$, otherwise 0.
Analogously for 
{\em geq(x,y)} (greater or equal),
{\em grt(x,y)} (greater).
Operand 
{\em and(x,y)} returns 1 if $x>0$ and $y>0$, otherwise 0.
Analogously for {\em or(x,y)}. 
Operand {\em not(x)} returns
1 if $x \leq 0$, otherwise 0.

\item {\sc Simple Stack Manipulators.}
{\em del()} decrements {\em dp};
{\em clear()} sets $dp :=0$; 
{\em dp2ds()} returns {\em dp};  
{\em setdp(x)} sets  $dp :=x$; 
{\em ip2ds()} returns $cs[cp].ip$; 
{\em base()} returns $cs[cp].base$;
{\em fromD()} returns $Ds[Dp]$;  
{\em toD()} pushes $ds[dp]$ onto {\em Ds};  
{\em delD()} decrements {\em Dp};
{\em topf()} returns the integer name of the most recent self-made function;
{\em intopf()} and {\em outopf()} return its number of requested inputs and
outputs, respectively;
{\em popf()} decrements {\em fnp}, returning its old value;
{\em xmn(m,n)} exchanges the $m$-th and the $n$-th
elements of {\em ds}, measured from the stack's top; 
{\em ex()} works like {\em  xmn(1,2)};
{\em xmnb(m,n)} exchanges the $m$-th and the $n$-th
elements {\em above} the current base $ds[cs[cp].base]$;
{\em outn(n)} returns  $ds[dp-n+1]$;
{\em outb(n)} returns  $ds[cs[cp].base + n]$ (the $n$-th
element above the base pointer);
{\em inn(n)} copies $ds[dp]$ onto $ds[dp-n+1]$;
{\em innb(n)} copies $ds[dp]$ onto $ds[cs[cp].base + n]$.

\item {\sc Pushing Code.} Instruction
{\em getq(n)} 
pushes onto {\em ds} the sequence beginning at the start address of
the $n$-th frozen program (either user-defined or frozen by {\sc oops}) 
and ending with the program's final token.  
{\em insq(n,a)} 
inserts the $n$-th frozen program above $ds[cs[cp].base+a]$,
then increases {\em dp} by the program size.
Useful for copying previously frozen code into modifiable
memory, to later edit  the copy.

\item {\sc Editing Strings on Stack.} Instruction
{\em cpn(n)} copies the n topmost {\em ds} entries onto the top of {\em ds},
increasing {\em dp} by $n$;
{\em up()} works like {\em  cpn(1)};
{\em cpnb(n)} copies $n$ {\em ds} entries above $ds[cs[cp].base]$ 
onto the top of {\em ds}, increasing {\em dp} by $n$;
{\em mvn(a,b,n)} copies
the $n$ {\em ds} entries starting with 
$ds[cs[cp].base+a]$ to 
$ds[cs[cp].base+b]$ and following cells, 
appropriately increasing {\em dp} if necessary;
{\em ins(a,b,n)} inserts the
$n$ {\em ds} entries above
$ds[cs[cp].base+a]$ after
$ds[cs[cp].base+b]$, appropriately increasing {\em dp};
{\em deln(a,n)} deletes
the $n$ {\em ds} entries above
$ds[cs[cp].base+a]$, 
appropriately decreasing {\em dp};
{\em find(x)}  returns the stack index of the topmost entry in {\em ds} matching $x$;
{\em findb(x)} the index of the first {\em ds} entry above base $ds[cs[cp].base]$ matching $x$.
Many of the above instructions can be used to edit stack contents
that may later be interpreted as executable code.

\end{enumerate}

\subsubsection{Control-Related Instructions}
\label{control}

Each call of callable code $f$ increments $cp$ and results in a
new topmost callstack entry. Functions to make and execute functions include:

\begin{enumerate}
\item 
Instruction {\em def(m,n)} 
defines a new integer function name (1 if it is the first,
otherwise the most recent name plus 1)  and
increments {\em fnp}. In the new {\em fns} entry we associate with the name: $m$ and $n$,
the function's expected numbers of input arguments and return values,
and the function's start address $cs[cp].ip + 1$ (right after the address
of the currently interpreted token {\em def}).

\item 
Instruction
{\em dof(f)}  calls $f$: it views $f$ as a function name,
looks up $f$'s address and input number $m$ and output number $n$,
increments $cp$,
lets $cs[cp].base$ point right below the $m$ topmost elements (arguments) in {\em ds}
(if $m < 0$ then $cs[cp].base=cs[cp-1].base$, that is, all
{\em ds} contents corresponding to the previous instance are viewed as arguments),
sets $cs[cp].out := n$,
and sets $cs[cp].ip$ equal to $f$'s address,
thus calling $f$.

\item 
{\em ret()} causes the current function call to return;
the sequence of the $n=cs[cp].out$
topmost values on {\em ds} is copied down such that it
starts in {\em ds} right above $ds[cs[cp].base]$, 
thus replacing the former input arguments;
{\em dp} is adjusted accordingly, 
and $cp$ decremented, thus transferring control
to the {\em ip} of the previous callstack entry
(no copying or {\em dp} change takes place if $n<0$ --- 
then we effectively
return the entire stack contents above $ds[cs[cp].base]$).
Instruction
{\em rt0(x)} calls {\em ret()} if $x \leq 0$ (conditional return).

\item 
{\em oldq(n)} 
calls the $n$-th frozen program (either user-defined or frozen by {\sc oops})
stored in $q$ below $a_{frozen}$, assuming (somewhat arbitrarily) zero inputs and outputs.

\item 
Instruction
{\em jmp1(val, n)} sets $cs[cp].ip$ equal to $n$ provided that $val$ exceeds zero 
(conditional jump, useful for iterative loops);
{\em pip(x)} sets $cs[cp].ip :=x$ (also useful
for defining iterative loops by manipulating the instruction pointer);
{\em bsjmp(n)} sets current instruction pointer $cs[cp].ip$ equal to the {\em address}
of $ds[cs[cp].base+n]$, thus interpreting stack contents above $ds[cs[cp].base+n]$
as code to be executed.

\item 
{\em bsf(n)} uses $cs$ in the usual way to {\em call}
the code starting at
$ds[cs[cp].base+n]$ (as usual, once the code is executed, we will return to the 
address of the next instruction right after {\em bsf});
{\em exec(n)} interprets $n$ as the number of an instruction and executes it.

\item 
{\em qot()} flips a binary flag {\em quoteflag} stored at address $a_{quoteflag}$ 
on tape as $z(a_{quoteflag})$. 
The semantics are: 
code in between two {\em qot}'s is quoted, not executed. More precisely,
instructions appearing 
between the $m$-th ($m$ odd) and the $m+1$st {\em qot} are not executed; 
instead their instruction numbers are sequentially pushed onto data stack {\em ds}.
Instruction {\em nop()} does nothing and may be used to structure programs.

\end{enumerate}
In the context of instructions such as {\em getq} and {\em bsf},
let us quote Koopman \cite{Koopman:89} (reprinted with friendly permission by
Philip J. Koopman Jr., 2002):
\begin{quote}
{\small 
{\em 
Another interesting proposal for stack machine program execution was
put forth by Tsukamoto (1977). He examined the conflicting virtues and
pitfalls of self-modifying code. While self-modifying code can be very
efficient, it is almost universally shunned by software professionals as
being too risky. Self-modifying code corrupts the contents of a program,
so that the programmer cannot count on an instruction generated by
the compiler or assembler being correct during the full course of a
program run.  Tsukamoto's idea allows the use of self-modifying code
without the pitfalls. He simply suggests using the run-time stack to
store modified program segments for execution. Code can be generated by
the application program and executed at run-time, yet does not corrupt
the program memory. When the code has been executed, it can be thrown
away by simply popping the stack.  Neither of these techniques is in 
common use today, but either one or both of them may eventually find an important application.
}
}
\end{quote}
Some of the instructions introduced above are almost exactly doing what has
been suggested by \cite{Tsukamoto:77}. Remarkably,
they turn out to be quite useful in the experiments (Section \ref{experiments}).

\subsubsection{Bias-Shifting Instructions to Modify Suffix Probabilities}
\label{self}

The concept of online-generated probabilistic programs 
with {\em ``self-referential''} instructions that modify the probabilities 
of instructions to be executed later was already implemented earlier by
\cite{Schmidhuber:97bias}.
Here we use the following primitives:
\begin{enumerate}

\item
{\em incQ(i)} 
increases the current probability of $Q_i$ by
incrementing $p[curp][i]$ and $sum[curp]$. 
Analogously for {\em decQ(i)} (decrement).
It is illegal to set all $Q$ probabilities 
(or all but one) to zero;
to keep at least two search options.  
{\em incQ(i)} and {\em decQ(i)} 
do not delete argument $i$ from {\em ds}, by not decreasing {\em dp}.

\item
{\em boostq(n)} sequentially goes through all instructions of 
the $n$-th self-discovered frozen program;
each time an instruction is recognized as some $Q_i$,
it gets {\bf boosted}: 
its numerator $p[curp][i]$ and the denominator $sum[curp]$ are increased
by $n_Q$.  (This is less specific than {\em incQ(i)}, but can be useful,
as seen in the experiments, Section \ref{experiments}.)

\item
{\em pushpat()} 
stores the current search pattern $pat[curp]$ 
by incrementing $patp$ and copying the sequence $pat[patp] :=  pat[curp]$;
{\em poppat()} decrements  $patp$,  returning its old value.
{\em setpat(x)} sets $curp := x$, thus defining the distribution
for the next search, given the current task. 

The idea is to give the system the opportunity to define 
several fairly arbitrary distributions on the possible
search options, and switch on useful ones when needed in a given computational
context, to implement conditional probabilities of
tokens, given a computational history.

\end{enumerate}
Of course, we could also {\em explicitly} implement tape-represented conditional
probabilities of tokens, given previous tokens or token sequences, 
using a tape-encoded, modifiable {\em probabilistic syntax diagram} for defining
modifiable {\em n-grams}. This may
facilitate the act of ignoring certain meaningless program prefixes
during search.  In the present implementation, however, the system has to
create / represent such conditional dependencies by
invoking appropriate subprograms including sequences of 
instructions such as {\em incQ()}, {\em pushpat()} and {\em setpat()}.

\subsection{Initial User-Defined Programs: Examples}
\label{user}

The user can declare initial, possibly recursive programs by composing the
tokens described above. 
Programs are sequentially written
into $q$, starting with $q_1$ at address 1.
To declare a new token (program) we write
{\em decl(m, n, {\sc name}, body)}, where
{\sc name} is the textual name of the code. Textual names are of interest
only for the user, since the system immediately translates any new name into
the smallest integer $> n_Q$ which gets associated with the topmost unused 
code address; then $n_Q$ is incremented.
Argument $m$ denotes the code's number of expected arguments
on top of the data stack {\em ds};
$n$ denotes the number of return values;
{\em body} is a string of names of previously 
defined instructions, and possibly one new name to allow
for cross-recursion.  Once the interpreter
comes across a user-defined token, it 
simply calls the code in $q$ starting with that body's first token; 
once the code is executed, the interpreter returns to the address of the next token,
using the callstack $cs$.
All of this is quite similar to the case of self-made functions defined by
the system itself --- compare instruction {\em def} in
section \ref{control}.

Here are some samples of user-defined tokens or programs composed from
the primitive instructions defined above.  Declarations
omit parantheses for argument lists of instructions. 

\begin{enumerate}
\item
{\em decl(0, 1, {\sc c999}, c5 c5 mul c5 c4 c2  mul mul mul dec ret)}
declares {\sc c999()}, a program without arguments, computing 
constant 999 and returning it on top of data stack {\em ds}.

\item
{\em 
decl(2, 1, {\sc testexp},    by2 by2 dec c3  by2 mul mul up  mul ret) 
}
declares {\sc testexp} {\em (x,y)}, which
pops two values $x,y$ from {\em ds} and returns $[6x (4y - 1)]^2$.

\item
{\em 
decl(1, 1, {\sc fac},        up c1 ex rt0    del up dec {\sc fac}  mul ret)
}
declares a recursive function {\sc fac}{\em (n)} which
returns 1 if $n=0$, otherwise returns $n \times$ {\sc fac}{\em(n-1)}.

\item
{\em 
decl(1, 1, {\sc fac2},       c1 c1 def up c1 ex rt0 del up   dec topf dof mul ret)   
}
declares {\sc fac2}{\em (n)}, which 
defines self-made recursive code 
functionally equivalent to {\sc fac}{\em (n)}, which calls itself by calling
the most recent self-made function even before it is completely defined.
That is, {\sc fac2}{\em (n)} not only computes {\sc fac}{\em (n)} 
but also makes a new {\sc fac}-like function.

\item
The following declarations are useful for defining and executing
recursive procedures (without return values) that expect as many inputs 
as currently found on stack {\em ds},
and call themselves on decreasing problem sizes. 
{\em defnp} first pushes onto auxiliary stack {\em Ds} the number of return values (namely, zero),
then measures the number $m$ of inputs on {\em ds} and pushes it onto {\em Ds},
then quotes (that is, pushes onto {\em ds}) the begin of the definition
of a procedure that returns if its topmost input $n$ is 0 and 
otherwise decrements  $n$. 
{\em callp} quotes a call of the most recently defined function / procedure. 
Both {\em defnp} and {\em callp} 
also quote code for making a fresh copy of the inputs of the most
recently defined code, expected on top of {\em ds}.
{\em endnp} quotes the code for returning, 
grabs from {\em Ds} the numbers of in and out values, 
and uses {\em bsf} to call the code generated so far on stack {\em ds} 
above the input parameters, applying this code (possibly located deep in $ds$)
to a copy of the inputs pushed onto the top of $ds$.

{\em 
decl(-1,-1,defnp,    c0 toD pushdp   dec toD qot def up rt0 dec intpf cpn qot   ret)      

decl(-1,-1,calltp,   qot topf dof    intpf   cpn qot ret) 

decl(-1,-1,endnp,qot ret qot fromD   cpnb fromD up   delD fromD ex bsf ret)
}

\item
Since our entire language is based on integer sequences, there is no obvious distinction
between data and programs. The following illustrative example demonstrates
that this makes functional programming  very easy:

{\em 
decl(-1, -1, {\sc tailrec},     qot c1 c1 def   up qot c2 outb  qot ex rt0 del
     up dec topf dof qot c3 outb qot ret qot c1 outb c3 bsjmp)
}
declares a tail recursion scheme {\sc tailrec} with a functional argument. 
Suppose the data stack {\em ds} holds three values $n$, {\em val}, and {\em codenum} 
above the current base pointer.
Then {\sc tailrec} will create a function that returns {\em val}  
if $n=0$, else applies 
the 2-argument function represented by {\em codenum}, where the arguments
are $n$ and the result of 
calling the 2-argument function itself on the value $n-1$. 

For example,
the following code fragment uses {\sc tailrec} to implement yet 
another version of {\sc fac}{\em (n)}:
{\em (qot c1 mul qot  {\sc tailrec} ret)}.
Assuming $n$ on {\em ds}, it first quotes the constant {\em c1} (the 
return value for the terminal case $n=0$) and 
the function {\em mul}, then applies {\sc tailrec}.

\end{enumerate}
The primitives of Section \ref{language} collectively embody a universal
programming language, computationally as powerful as 
the one of \cite{Goedel:31} or {\sc Forth} or {\sc Ada} or C.
In fact, a small fraction of the primitives would already 
be sufficient to achive this universality.
Higher-level programming languages can be incrementally
built based on the initial low-level {\sc Forth}-like language.

To fully understand a given program, one may need to know 
which instruction has got which number.
For the sake of completeness, and to permit precise
re-implementation, we include the full list 
here:

{\em 
1: 1toD,  2: 2toD,  3: mvdsk,  4: xAD,  5: xSA,  
6: bsf,  7: boostq,  8: add,  9: mul,  10: powr,  11: sub,
12: div,  13: inc,  14: dec,  15: by2,  16: getq,
17: insq,  18: findb, 19: incQ,  20: decQ,  21: pupat,
22: setpat,  23: insn,  24: mvn,  25: deln,  26: intpf,
27: def,  28: topf,  29: dof,  30: oldf,  31: bsjmp,
32: ret,  33: rt0,  34: neg,  35: eq,  36: grt,  37: clear,  
38: del,  39: up,  40: ex,  41: jmp1,  42:
outn,  43: inn,  44: cpn,  45: xmn,  46: outb,  47: inb,
48: cpnb,  49: xmnb,  50: ip2ds,  51: pip,  52: pushdp,
53: dp2ds, 54: toD,  55: fromD,  56: delD,  57: tsk,
58: c0,  59: c1,  60: c2,  61: c3,  62: c4,  63: c5,
64: exec,  65: qot,  66: nop,  67: fak,  68: fak2,  69: c999,  
70: testexp,  71: defnp,  72: calltp,  73: endnp.
}

\bibliography{bib}
\end{document}